\documentclass[preprint,5p]{elsarticle}

\usepackage{graphicx}
\usepackage{amsmath,amssymb,amsfonts}
\usepackage{algpseudocode,algorithm}
\usepackage{subcaption}
\usepackage{makecell}
\usepackage{url}
\usepackage{multirow}

\usepackage{pifont}
\newcommand{\cmark}{\ding{51}}
\newcommand{\xmark}{\ding{55}}

\journal{Robotics and Autonomous Systems}

\begin{document}

\sloppy

\begin{frontmatter}



\title{LiRA: Light-Robust Adversary for \\Model-based Reinforcement Learning in Real World}


\author{Taisuke Kobayashi\corref{cor}}
\ead{kobayashi@nii.ac.jp}
\ead[url]{https://prinlab.org/en/}

\cortext[cor]{Corresponding author}

\address{National Institute of Informatics (NII) and The Graduate University for Advanced Studies (SOKENDAI), Tokyo, Japan}

\begin{abstract}

Model-based reinforcement learning has attracted much attention due to its high sample efficiency and is expected to be applied to real-world robotic applications.
In the real world, as unobservable disturbances can lead to unexpected situations, robot policies should be taken to improve not only control performance but also robustness.
Adversarial learning is an effective way to improve robustness, but excessive adversary would increase the risk of malfunction, and make the control performance too conservative.
Therefore, this study addresses a new adversarial learning framework to make reinforcement learning robust moderately and not conservative too much.
To this end, the adversarial learning is first rederived with variational inference.
In addition, \textit{light robustness}, which allows for maximizing robustness within an acceptable performance degradation, is utilized as a constraint.
As a result, the proposed framework, so-called LiRA, can automatically adjust adversary level, balancing robustness and conservativeness.
The expected behaviors of LiRA are confirmed in numerical simulations.
In addition, LiRA succeeds in learning a force-reactive gait control of a quadrupedal robot only with real-world data collected less than two hours.

\end{abstract}


\end{frontmatter}


\section{Introduction}

Reinforcement learning (RL) attracts increasing attention as methodology for robots to learn stochastic control policies that enable them to accomplish a given task by trial and error~\citep{sutton2018reinforcement}.
In particular, model-based RL has high expectations for real-world robotic applications due to its excellent sample efficiency~\citep{chua2018deep,williams2018information}, and various robotic applications have been reported recently: e.g.
manipulation on complex dynamics of dexterous hand~\citep{nagabandi2020deep}, cloth~\citep{luque2024model}, and even grinding of object~\citep{hachimine2023learning};
locomotion by quadruped~\citep{yang2020data}, biped~\citep{kuo2023reinforcement}, and quadrotor~\citep{lambert2019low};
and their combination~\citep{arcari2023bayesian}.
Although the framework proposed in this study can be applied to model-free RL, for the sake of simplicity, this paper limits its focus to model-based RL.

In addition to the sample efficiency (and, of course, control performance), recent RL studies have often aim to maximize robustness.
This would enable robots to cope with unexpected situations in a faced environment, which can easily be caused by a variety of factors (e.g. biases in empirical data and unobservable information).
Robustness is essential for robots to work in the real world ``reliably'' because it is difficult to observe all information there~\citep{zhang2021reinforcement}.

However, achieving robustness is not a free lunch, and in the context of RL, it often requires to experience a wide variety of situations, making its sample efficiency poor~\citep{morimoto2005robust}.
While the recent progresses have mitigate this problem by using the fast and parallel simulatior~\citep{andrychowicz2020learning} and/or more flexible adversarial learning~\citep{pinto2017robust}, the side effect of maximizing robustness, i.e. conservative policies that degrade control performance under normal conditions, is raised as a new problem~\citep{lechner2023revisiting}.
Because this tradeoff is inherently unavoidable, another approach has been proposed in which, instead of increasing robustness, uncertainty is revealed online and reflected in models and/or policies~\citep{ilboudo2023domains}.
In this case, the risk of erroneous uncertainty estimation is caused, and the computation cost during inference is increased due to estimating uncertainty, which affects real-time control.
Although the details of these problems in related work are summarized in the next section, to my knowledge, no framework or theoretical system has yet been proposed that balances all of these issues.

\begin{figure*}[tb]
    \centering
    \includegraphics[keepaspectratio=true,width=0.96\linewidth]{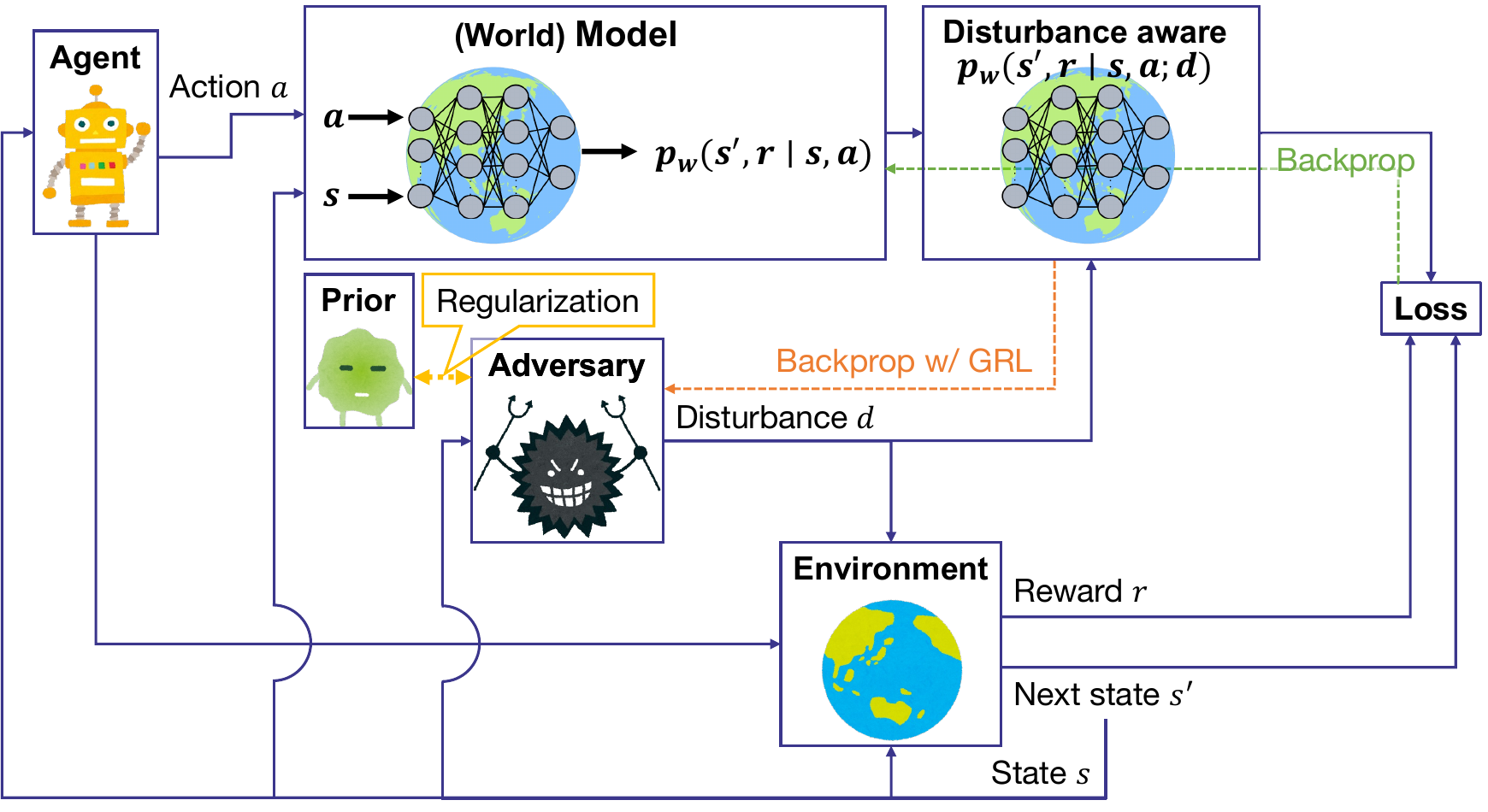}
    \caption{Proposed framework, LiRA:
    a trainable adversary generates a disturbance to deteriorate the predictive performance of (world) model in an adversarial manner;
    by limiting the deterioration of the predictive performances between disturbance-marginalized and disturbance-aware models, the adversary level is automatically tuned, reverting to its prior.
    }
    \label{fig:lira}
\end{figure*}

In line with this background, this study proposes a new framework, so-called LiRA (Light-Robust Adversary) (see Fig.~\ref{fig:lira}), for adversarial learning that satisfies the following two real-world requirements:
\begin{enumerate}
    \item During learning, the robustness is moderately and safely improved while mitigating learning collapse and policy conservativeness;
    \item During inference, the computational cost for selecting the optimal action does not increase.
\end{enumerate}
To this end, adversarial learning is first rederived according to variational inference~\citep{kingma2014auto}.
Then, as a new definition of robustness, \textit{light robustness}~\citep{fischetti2009light} is integrated with it.
This relaxes full robustness by imposing an inequality constraint that limits the degradation from ideal performance due to disturbances within a specified threshold (this corresponds to the conservativeness).
Through the Lagrangian method~\citep{kobayashi2023soft}, this inequality constraint is transformed into a loss function for numerical optimization, allowing for automatic tuning of the adversary level.
In addition, as such a learning process makes the model marginalized w.r.t. disturbances moderately robust, the disturbances can remain unknown at inference time, requiring no estimation cost or additional conditioning.

The behaviors of the proposed LiRA are confirmed by numerical simulations using Mujoco~\citep{todorov2012mujoco}.
LiRA is not too conservative in control performance only with the nominal noise, and the degradation of control performance in response to disturbance intensity is suppressed.
In addition, according to the specified conservativeness, the proposed LiRA attempts to increase the robustness by increasing the adversary level for the condition with weaker disturbance sensitivity, and vice versa.
Such an auto-tuning capability also yields self-paced curriculum learning~\citep{kumar2010self,klink2020self}, where disturbances are suppressed during the under-performance phase and reinforced as learning progresses.
These results indicate that LiRA can achieve a good balance between robustness and conservativeness.

Finally, a real-robot learning using LiRA is demonstrated without simulations.
By applying virtual external forces to its internal impedance model, a quadrupedal robot perceives disturbances only during learning.
This task can be performed by LiRA only on the real robot with appropriate suppression of excessive disturbance, whereas a fully adversarial learning has a high risk of malfunctioning the robot.
Although learning with nominal noise alone should not be sufficient to generate a force-responsive gait, LiRA makes it possible by moderately experiencing various situations during learning.
Thus, this demonstration illustrates that LiRA can induce the robot to acquire the desired behaviors only with the limited data collected from the real world in short time (less than two hours in this demonstration).

\begin{table*}[tb]
    \caption{Requirements for (model-based) RL policies}
    \label{tab:related_work}
    \centering
    \begin{tabular}{l cccc}
        \hline\hline
        Method & Less conservative & More robust & More efficient (at learning) & More efficient (at inference)
        \\
        \hline
        Vanilla
        & \cmark\cmark
        & \xmark
        & \cmark\cmark
        & \cmark
        \\
        Domain randomization
        & \xmark
        & \cmark\cmark
        & \xmark
        & \cmark
        \\
        Adversarial learning
        & \xmark
        & \cmark\cmark
        & \cmark
        & \cmark
        \\
        Domain adaptation
        & \cmark\cmark
        & \cmark
        & \xmark
        & \xmark
        \\
        LiRA (proposal)
        & \cmark
        & \cmark
        & \cmark
        & \cmark
        \\
        \hline\hline
    \end{tabular}
\end{table*}

The contributions of this paper are threefold.
\begin{enumerate}
    \item Theoretical aspect: By integrating the variational inference and light robustness, a novel learning rule, which provides the moderate robustness, is theoretically derived.
    \item Implementation aspect: Three modules are introduced for the learning rule to function as expected, and their necessity is verified by simulations.
    \item Practical aspect: By automatically adjusting the state-dependent adversary level, the efficient and safe adversarial learning is demonstrated in the real world, acquiring the robust policy.
\end{enumerate}
Robustness of policies that do not impair control performance is critical in real-world applications, but its realization so far is based on the use of simulations that differ from reality in terms of efficiency and safety~\citep{hoeller2024anymal,radosavovic2024real}.
On the other hand, this research, which guarantees efficiency and safety by using adversarial learning without excessive attacks, makes it possible to obtain robust policies even only in the real world.
In other words, the proposed LiRA can be applied to problems that are difficult to simulate or take longer to simulate than in real time, and it can be adapted to real-world robotic systems that change in a non-stationary manner.

\subsection{Related work}

To enhance the robustness of RL policies, a naive approach is domain randomization on simulations~\citep{tobin2017domain,ramos2019bayessim,muratore2022neural}, which encompasses a variety of optimal policies by learning from data experienced in simulations driven by various simulation parameters (friction, mass, texture, and so on).
Although this approach has had much success in recent years~\citep{hoeller2024anymal,radosavovic2024real}, it should be inefficient unless data are collected in parallel using plenty of computational resources, which waste huge amounts of energy~\citep{andrychowicz2020learning,rudin2022learning}.
It is therefore difficult for real-world robots to continuously learn on-board with this approach.
In addition, this approach implicitly assumes that the real world can be represented by either of the randomly selected simulation parameters~\citep{chen2022understanding}, which requires either deep domain knowledge to find an appropriate parameter space or a broader parameter space to be tried.
The former lacks generality, and the latter further increases the conservativeness of policies.

As a more efficient approach, an adversarial learning can be considered, in which an adversary is introduced that actively interferes with the robot's task accomplishment~\citep{pinto2017robust,gleave2020adversarial,zhai2022robust}.
Such active disturbances (modeled by highly expressive neural networks) can produce complex and challenging experiences more efficiently than random ones.
However, this approach inevitably inherits the mode collapse issue in adversarial learning~\citep{srivastava2017veegan,liu2019spectral}, making learning the optimal policy unstable.
In addition, in real-world robots, excessive disturbances might damage themselves and/or objects in the faced environment.
The previous study that mitigates these issues is RRL-Stack~\citep{huang2022robust}, which is based on Stackelberg game and would be the most similar to the proposed method, but it requires additional rollouts with oracle policies, sacrificing the sample efficiency.
Furthermore, one of the adverse effects of maximizing robustness (caused in both of the above) is to make the policy too conservative~\citep{petrik2019beyond,lechner2023revisiting,huang2023trade}.
In other words, since robots have to take into account contingencies that may not occur in practice, they tend to choose safer and more secure actions.
If the maximum disturbance intensity could appropriately be designed in advance, such conservativeness would be controlled at the minimum.
However, as the relationship between the policy performance and the disturbance intensity is nonlinear and varies from situation to situation, this solution is infeasible.

This issue would be mitigated by domain adaptation~\citep{yu2017preparing,semage2022uncertainty,ilboudo2023domains}, which estimates the uncertainty in the faced environment (e.g. represented as simulation parameters and disturbances) and incorporates them into the policy.
As long as the uncertainty estimator is accurate, this approach can acquire the necessary and sufficient balance between robustness and conservativeness.
However, its learning cost should be equal to or larger than that of domain randomization, since the effects of any potential uncertainty must be learned in advance.
In addition, the extra computational cost of the uncertainty estimator and the uncertainty-conditioned policy can easily be an obstacle to real-world robot control, which is strictly time-constrained.
In particular, it has been reported that in the model-based RL algorithms, where model predictive control (MPC)~\citep{williams2018information,kobayashi2022real} is employed in combination, reducing the extra computational cost as much as possible leads to better control performance because the MPC optimization converges in time~\citep{lenz2015deepmpc,kobayashi2023sparse}.
As the estimation error would also be raised as our concern, an alternative solution to this domain adaptation should be considered.

Thus, each of the conventional approaches has different drawbacks, as summarized in Table~\ref{tab:related_work}.
Since all of these features are necessary for real-world applications, LiRA aims to achieve a good balance between all of them.

\section{Preliminaries}

\subsection{Model-based reinforcement learning}

In RL~\citep{sutton2018reinforcement}, an agent aims to gain the maximum rewards from an environment in the future.
For mathematical formulation of the relationship between the agent and environment, Markov decision process (MDP) is basically assumed, i.e. $(\mathcal{S}, \mathcal{A}, p_e, r)$.
Here, $\mathcal{S}$ is the state space, $\mathcal{A}$ is the action space, $p_e: \mathcal{S} \times \mathcal{A} \to \mathcal{S}$ is the state transition probability (a.k.a. dynamics), and $r: \mathcal{S} \times \mathcal{A} \times \mathcal{S} \to \mathcal{R}$ is the reward function to evaluate each transition.
Here, $\mathcal{R} \subseteq \mathbb{R}$ holds normally, but the case of multi-objective RL, where multiple rewards are simultaneously considered~\citep{hayes2022practical}, is assumed for generality.
That is, the output of reward function is vectorized as $\mathcal{R} \subseteq \ \mathbb{R}^{|\mathcal{R}|}$.

Under MDP, this study solves the following optimization problem for acquiring the optimal (stochastic) policy $\pi^\ast: \mathcal{S} \to \mathcal{A}$, which should be able to accomplish the task defined by $r$, at the discrete time step $t \in \mathbb{N}$.
\begin{align}
    &\pi^\ast(a_t \mid s_t) = \arg\max_{\pi(a_t \mid s_t)} \sum_{k=0}^H u(r_{t+k})
    \label{eq:def_rl} \\
    \mathrm{s.t.} &\begin{cases}
        r_{t+k} &= r(s_{t+k}, a_{t+k}, s^\prime_{t+k})
        \\
        s^\prime_{t+k} & \sim p_e(s_{t+k+1} \mid s_{t+k}, a_{t+k})
        \\
        a_{t+k} & \sim \pi(a_{t+k} \mid s_{t+k})
    \end{cases}
    \nonumber
\end{align}
where $u: \mathcal{R} \to \mathbb{R}$ denotes the utility function to make rewards scalar (the simple summation, in this paper).
$H \in \mathbb{N}$ denotes the horizon indicating how far into the future to be considered.

If the agent knows $p_e$ and $r$ accurately, this problem can numerically be solved using MPC (in this paper, AccelMPPI~\citep{kobayashi2022real} is employed).
Therefore, model-based RL algorithms explicitly approximate them as a model ($p_w$ in this paper) using, for example, deep neural networks with parameters $\theta$~\citep{chua2018deep,kobayashi2023sparse}.
\begin{align}
    &\theta^\ast = \arg\min_{\theta} \mathbb{E}_{p_e, r, \pi}[- \ln p_w(s^\prime, r \mid s, a; \theta)]
    \label{eq:model_vanilla}\\
    \mathrm{s.t.} &\begin{cases}
        r &= r(s, a, s^\prime)
        \\
        s^\prime &\sim p_e(s^\prime \mid s, a)
        \\
        a &\sim \pi(a \mid s)
    \end{cases}
    \nonumber
\end{align}
Note that, in this paper, $r$ is alternatively represented as a probability $p_r$ conditioned on $s$ and $a$ except $s^\prime$ for simplicity.
The obtained $p_w$ replaces $p_r$ and $r$ in eq.~\eqref{eq:def_rl}.

\subsection{Adversarial learning}

To make $\pi^\ast$ obtained through the above optimization problem robust, worst-case scenarios for the objective function (i.e. the sum of predicted reward, called the return) are basically considered~\citep{kohler2020computationally,zanon2020safe}.
However, this incurs an extra computational cost at inference time compared to the case simply using expected value of the prediction.
In addition, using these methods, it is difficult to determine whether robustness is due to the model or to such a robust MPC.
This study therefore focuses on the direction that robust control is achieved by making $p_w$ robust during learning, referring to the result reported in the literature~\citep{aotani2024cooperative} that control performance becomes implicitly robust if $p_w$ is optimized in consideration of events that are rare in reality.

To make $p_w$ fully robust, it is effective to intentionally allow the agent to experience the rare events.
That is, this purpose can be achieved by adversarial learning with the following min-max problem, instead of eq.~\eqref{eq:model_vanilla}:
\begin{align}
    &\theta^\ast, \phi^\ast = \arg\min_{\theta}\max_{\phi} \mathbb{E}_{\tilde{p}_e, \tilde{r}, \pi, \varpi}[- \ln p_w(s^\prime, r \mid s, a; \theta)]
    \label{eq:model_robust}\\
    \mathrm{s.t.} &\begin{cases}
        r &= \tilde{r}(s, a, s^\prime; d)
        \\
        s^\prime &\sim \tilde{p}_e(s^\prime \mid s, a; d)
        \\
        a &\sim \pi(a \mid s)
        \\
        d &\sim \varpi(d \mid s; \phi)
    \end{cases}
    \nonumber
\end{align}
where $\varpi: \mathcal{S} \to \mathcal{D}$ denotes the learnable adversary (or, disturbance generator) with parameters $\phi$.
Note that the disturbance $d$ should have upper and lower bounds and be distributed around zero to avoid the collapse and bias of the original environment.
Therefore, it is assumed that $d \in \mathcal{D} = [-d^\mathrm{max}, d^\mathrm{max}]$ with $d^\mathrm{max}$ the maximum disturbance intensity, which is specified in advance as a hyperparameter.

As a remark, $\varpi$ can be conditioned on $a$ in addition to $s$, but in that case, the effects of $d$ are too strong because $d$ can easily interfere with $a$ by determining later than $a$.
As this study is in favor of moderate robustness, $a$ was omitted from the condition for $\varpi$.
In addition, the way $d$ acts on the environment is a discussion for robotic applications, so it was also omitted in this paper in favor of theory.

\section{LiRA: Light-robust adversary}

\subsection{Overview}

The basic adversarial learning suffers from the mode collapse issue~\citep{srivastava2017veegan,liu2019spectral}, where the min-max problem becomes unbalanced (in most cases, the adversary becomes dominant), preventing the model from robust learning properly.
In order to improve the robustness of the control performance to the extent that it does not become too conservative, this paper proposes LiRA.
LiRA utilizes two theoretical tricks in its formulation.

First, the adversarial learning in eq.~\eqref{eq:model_robust} is reformulated by variational inference~\citep{kingma2014auto}, introducing the necessary components in a natural way.
In particular, the function to regularize the adversary to its prior distribution is derived, thus inhibiting excessive attacks by the adversary, albeit passively.
Then, the light robustness~\citep{fischetti2009light} is applied as an additional constraint for the gap of predictive performance of models with/without disturbance information, which can be converted to a regularization term according to Lagrangian method with an auto-tuned weight parameter.
This parameter has a role of adjusting adversary level automatically, actively balancing the robustness and conservativeness.

To appropriately solve the derived optimization problem, this paper implements three practical tricks (see the section~\ref{sec:impl}).
First, noting that the bias due to disturbances should ideally be zero, a design constraint is provided by a restricted normalizing flows (RNF)~\citep{kobayashi2023design} such that the expected values of the two models are consistent with each other.
Second, to make a gradient reversal layer (GRL)~\citep{ganin2016domain} available for the stable and efficient adversarial learning, a trick to hindsight the computational graph from the disturbance generator to the model via the disturbance, so-called hindsight reparameterization gradient (HRG), is newly developed.
Third, a new equality constraint to actively reflect the worst cases for either the model or the disturbance generator, midrange-mean balancing (MMB), is given, following the literature~\citep{aotani2024cooperative}, which efficiently improves the robustness of the model by (moderately) prioritizing the worst-case data for the performance.

\begin{figure}[tb]
    \centering
    \includegraphics[keepaspectratio=true,width=0.96\linewidth]{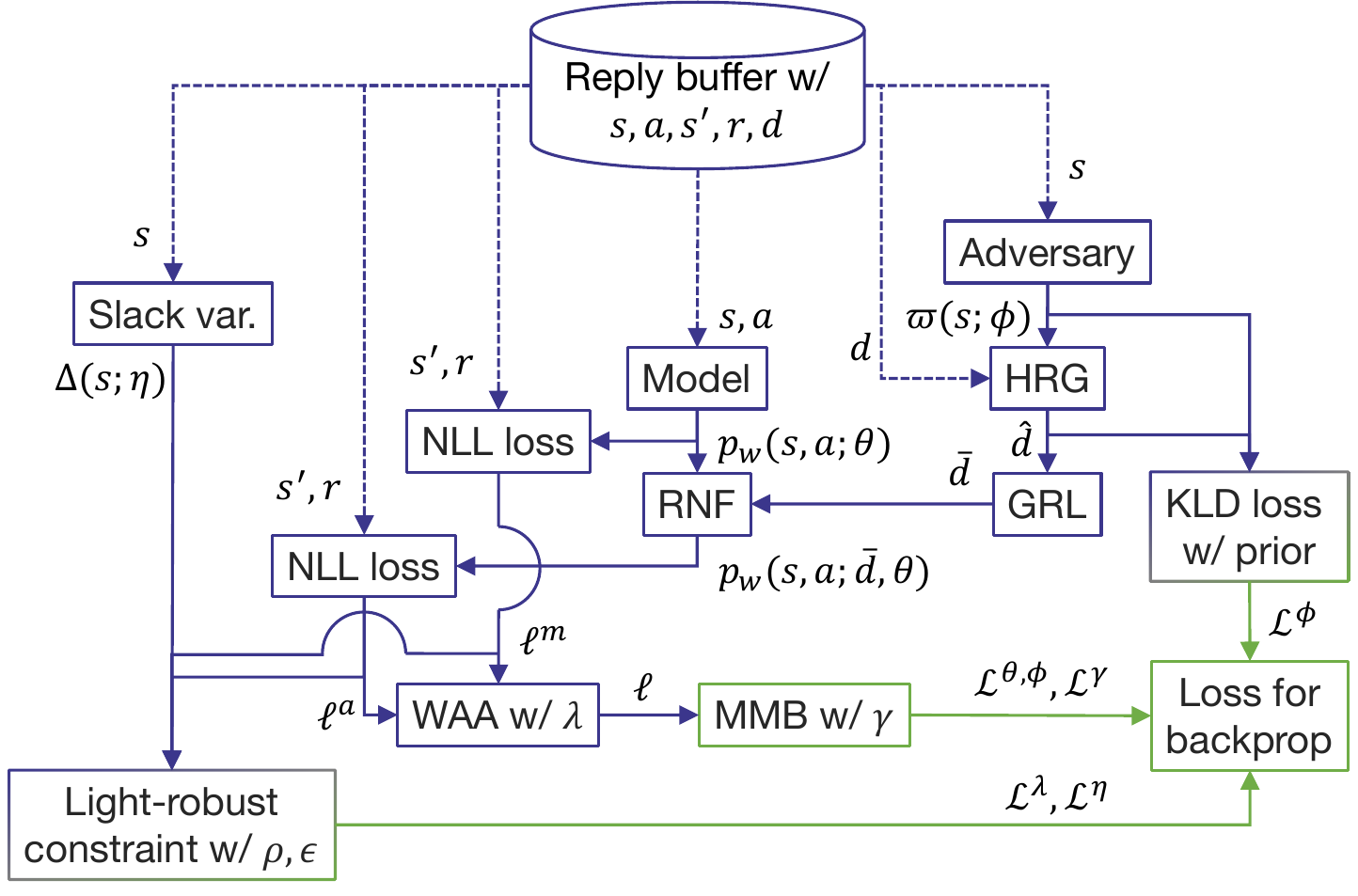}
    \caption{Process to calculate loss of LiRA:
    the dashed lines show the non-backpropagatable signals replayed from the buffer;
    the blue lines and blocks show per-sample processing, while the green lines and blocks are for processing for the batch, making values of samples a scalar loss.
    }
    \label{fig:pipeline}
\end{figure}

The details of the above proposal will be presented in the following sections.
Especially, its process for computing the loss function to be minimized is summarized in Fig.~\ref{fig:pipeline}.
The overall pseudocode of LiRA is also shown in Alg.~\ref{alg:lira}.
Note that the loss to be minimized in this pseudocode, $\mathcal{L}^x$, has the computational graph only for $x$ and cuts for the other variables to be optimized.
In addition, $\mathcal{L}^\phi$ corresponds to Kullback-Leibler divergence approximated by Monte Carlo method, which is necessary as the adversary is designed by normalizing flows.

\begin{algorithm}[tb]
    \caption{Pseudocode of LiRA}
    \label{alg:lira}
    \begin{algorithmic}[1]
        \State{Initialize network parameters $\theta$, $\phi$, $\eta$}
        \State{Initialize auto-tuned parameters $\lambda \simeq 1/2$, $\gamma \simeq 0$}
        \State{Initialize replay buffer $B = \emptyset$}
        \While{not converged}
            \\\hrulefill\Comment{Data collection}
            \While{not terminated or truncated}
                \State{Get the current state $s$}
                \State{Solve the action $a \sim \pi(a \mid s; \theta)$ by MPC}
                \State{Set the disturbance $d \sim \varpi(d \mid s; \phi)$}
                \State{Get the next state $s^\prime$ and reward $r$}
                \State{Store experience $B = B \cup (s, a, s^\prime, r, d)$}
            \EndWhile
            \\\hrulefill\Comment{Learning}
            \For{$\{ (s_i, a_i, s^\prime_i, r_i, d_i) \}_{i=1}^N \subset B$}
                \State{$\hat{d}_i = \mathrm{HRG}(\varpi(d \mid s_i; \phi), d_i)$ in eq.~\eqref{eq:hrg}}
                \State{$\bar{d}_i = \mathrm{GRL}(\hat{d}_i)$ in~\citep{ganin2016domain}}
                \State{$\Delta_i = \Delta(s_i; \eta)$}
                \State{$\ell^m_i = - \ln p_w(s^\prime_i, r_i \mid s_i, a_i; \theta)$}
                \State{$\ell^a_i = - \ln p_w(s^\prime_i, r_i \mid s_i, a_i; \bar{d}_i, \theta)$ w/ RNF~\citep{kobayashi2023design}}
                \State{$\ell_i = \lambda \ell^m_i + (1 - \lambda) \ell^a_i$}
                \State{$\mathcal{L}^{\theta,\phi}, \mathcal{L}^\gamma = \mathrm{MMB}(\{\ell_i\}_{i=1}^N; \gamma)$ in eq.~\eqref{eq:mmb}}
                \State{$\mathcal{L}^\phi = \frac{\beta}{N} \sum_i \ln \varpi(\hat{d}_i \mid s_i; \phi) - \ln \varpi(\hat{d}_i)$}
                \State{$\mathcal{L}^\lambda = - \frac{1}{N} \sum_i \lambda \delta(s^\prime_i, r_i, s_i, a_i, d_i)$}
                \State{$\mathcal{L}^\eta = \frac{1}{N} \sum_i \ell^\Delta(s^\prime_i, r_i, s_i, a_i, d_i)$ in eq.~\eqref{eq:slack}}
                \State{Minimize $\mathcal{L}^{\theta,\phi} + \mathcal{L}^\phi + \mathcal{L}^\lambda + \mathcal{L}^\eta + \mathcal{L}^\gamma$}
            \EndFor
        \EndWhile
    \end{algorithmic}
\end{algorithm}

\subsection{Adversarial learning with variational inference}

First, adversarial learning defined in eq.~\eqref{eq:model_robust} is redefined based on variational inference as a basis for incorporating the light robustness introduced in the next section.
Specifically, we can focus on the fact that the disturbance $d$ is usually unobservable and regarded to be the latent variable.
In maximizing the log-likelihood of disturbance-marginalized model, $p_w(s^\prime, r \mid s, a; \theta)$, in which the unobservable $d$ is excluded by marginalization, maximizing its worst case is more appropriate from the point of view of robustness.
That is, by introducing the disturbance-aware model and the prior (a.k.a. nominal noise), the following evidence lower bound is derived according to Jensen's inequality (and inequality between expected value and infimum).
\begin{align}
    & \ln p_w(s^\prime, r \mid s, a; \theta)
    \nonumber \\
    =& \ln \mathbb{E}_{\varpi(d)}[p_w(s^\prime, r \mid s, a; d, \theta)]
    \nonumber \\
    \geq& \mathbb{E}_{\varpi(d \mid s)}[\ln p_w(s^\prime, r \mid s, a; d, \theta)] - \mathrm{KL}(\varpi(d \mid s; \phi) || \varpi(d))
    \nonumber \\
    \geq& \inf_{\varpi(d \mid s; \phi)}[\ln p_w(s^\prime, r \mid s, a; d, \theta)] - \mathrm{KL}(\varpi(d \mid s; \phi) || \varpi(d))
    \label{eq:model_vi}
\end{align}
where $\mathrm{KL}(\cdot || \cdot)$ denotes Kullback-Leibler divergence between two probabilities.
When maximizing this lower bound, the adversary tries to minimize $\ln p_w(s^\prime, r \mid s, a; d, \theta)$ to take into account the rare events, while regularizing $\varpi(d \ mid s; \phi) \to \varpi(d)$.
That is, the min-max problem in eq.~\eqref{eq:model_robust} seems to be extended by adding the regularization, which might alleviate the mode collapse with the excessive attacks by the adversary.

The degree of regularization is commonly adjusted by introducing the gain $\beta \geq 0$~\citep{higgins2017beta} ($\beta=0$ means no regularization as like eq.~\eqref{eq:model_robust}).
Although this regularization is expected to reduce the generation of excessive disturbances, how it ($\beta$, more specifically) affects to the balance between robustness and conservativeness is unclear.
For developing an auto-tuning mechanism of $\beta$ (or other alternative gain) to achieve the desired balance, some kind of criteria that can be intuitively specified by the user are needed.

In addition, we need to focus on the fact that the newly introduced disturbance-aware model, $p_w(s^\prime, r \mid s, a; d, \theta)$, cannot be used when $d$ is unknown, so it is used only for adversarial learning.
Only if an additional disturbance estimator could be introduced, it could be used even during inference to enable the domain adaptation.
In that light, as $p_w(s^\prime, r \mid s, a; \theta)$ is not included in the above lower bound, it should be additionally optimized by explicitly considering some kind of conditions.

\subsection{Integration with light robustness}

To address these remaining issues, the \textit{light robustness}~\citep{fischetti2009light} is integrated with the maximization problem of the above lower bound.
The light robustness establishes a tolerance for performance degradation due to disturbances and assigns that constraint to the optimization problem.
This allows for a more intuitive setting since the ``relative'' tolerance can be defined, in comparison to, for example, the degree of regularization to the prior and the absolute predictive performance of the model.

Specifically, for any state (and action), the following constraint is applied.
\begin{align}
    &- \ln p_w(s^\prime, r \mid s, a; \theta) \leq - \ln p_w(s^\prime, r \mid s, a; d, \theta) + \rho
    \nonumber \\
    &\underbrace{\ln p_w(s^\prime, r \mid s, a; d, \theta) - \ln p_w(s^\prime, r \mid s, a; \theta) - \rho + \Delta(s; \eta)}_{\delta(s^\prime, r, s, a, d)}
    \nonumber \\
    &= 0
    \label{eq:light_robust}
\end{align}
where, $\rho \geq 0$ denotes the tolerance of performance degradation and $\Delta \geq 0$ denotes the slack variable, which represents the different between the left and right sides and is approximated by parameters $\eta$.
The first line is the inequality according to the original light robustness, and the second line is a clarification for this study.
Note that although it may be temporarily violated depending on the initialization of the models, $\ln p_w(s^\prime, r \mid s, a; d, \theta) \leq \ln p_w(s^\prime, r \mid s, a; \theta)$ holds in most cases because the likelihood is higher when conditioned with more necessary information.

This constraint can be converted into the corresponding regularization term via Lagrangian method with an auto-tuned gain $\lambda$, as shown in the literature~\citep{kobayashi2023soft}.
In summary, the proposed LiRA solves the following optimization problem to suppress conservativeness while making the model moderately robust.
\begin{align}
    &\theta^\ast, \phi^\ast = \arg\min_{\theta}\max_{\phi} \mathbb{E}_{\tilde{p}_e, \tilde{r}, \pi, \varpi}
    [- \lambda \ln p_w(s^\prime, r \mid s, a; \theta)
    \nonumber \\
    &\quad\quad\quad\quad\quad\quad\quad\quad\quad\ - (1 - \lambda) \ln p_w(s^\prime, r \mid s, a; d, \theta)
    \nonumber \\
    &\quad\quad\quad\quad\quad\quad\quad\quad\quad\ - \beta \mathrm{KL}(\varpi(d \mid s; \phi) || \varpi(d))]
    \label{eq:model_lira}\\
    \mathrm{s.t.} &\begin{cases}
        r &= \tilde{r}(s, a, s^\prime; d)
        \\
        s^\prime &\sim \tilde{p}_e(s^\prime \mid s, a; d)
        \\
        a &\sim \pi(a \mid s)
        \\
        d &\sim \varpi(d \mid s; \phi)
    \end{cases}
    \nonumber
\end{align}
where, $\lambda$ (and $\eta$ for approximating $\Delta$) can be optimized, as described in \ref{app:lambda}.
Note that $\lambda$ is a Lagrange multiplier, so $\lambda \in \mathbb{R}$ holds in general, but if $\lambda < 0$ and $\lambda > 1$, $\theta$ will be learned in a direction that degrates the predictive performance of the models.
Since this is not in line with the original purpose of model learning, $\lambda$ is restricted within $[0, 1]$ in this study.

If the disturbance is too strong, $\delta > 0$ is likely to occur, leading to $\lambda \to 1$.
As a result, the adversarial learning for the disturbance-aware model is suppressed, and the adversary is dominantly regularized to its prior, weakening the disturbance.
On the other hand, if the disturbance is too weak, $\delta < 0$ and $\lambda \to 0$ are expected, and the adversarial learning is activated to strengthen the disturbance.
In this way, LiRA automatically adjusts the disturbance intensity to be moderately robust and not exceed the specified tolerance (a.k.a. conservativeness).

Note that $\lambda$ is given as a scalar for simplicity, but since the light-robust constraint is given for each state, $\lambda$ can be a function of $s$ with parameters $\zeta$.
In this way, the state-dependent adversary level can be expected.
In other words, it might behave in such a way that $\lambda$ is increased when the agent is in the state sensitive to the disturbance intensity and $\lambda$ is decreased when the state is relatively safe.

\section{Implementation tricks}
\label{sec:impl}

\subsection{Restricted normalizing flows (RNF)}

\begin{figure}[tb]
    \centering
    \includegraphics[keepaspectratio=true,width=0.96\linewidth]{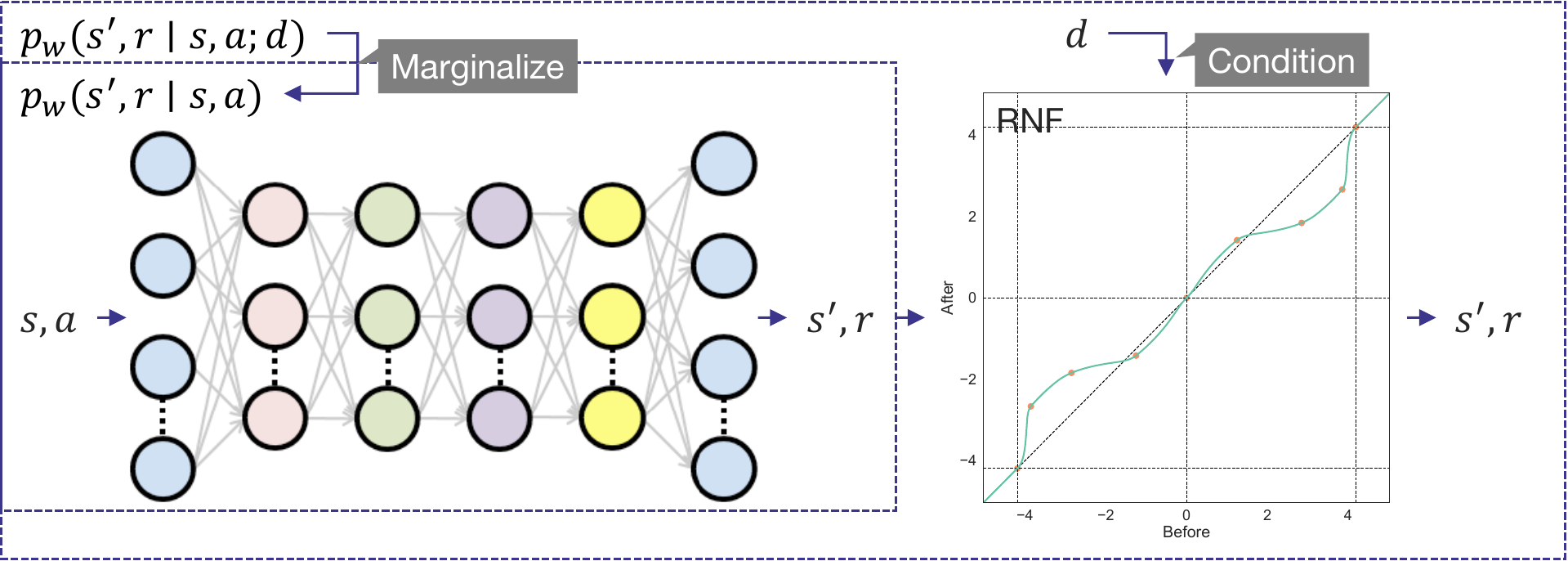}
    \caption{Restricted normalizing flows (RNF):
    when the base distribution is limited to be symmetric and the (conditional) invertible transformations are only odd functions, the mean (i.e. the center of distribution) is carried over to the converted distribution.
    }
    \label{fig:rnf}
\end{figure}

In LiRA, the disturbance-aware/marginalized models are learned independently in order to compare their prediction performance.
By using conditional normalizing flows (CNF)~\citep{papamakarios2021normalizing}, the disturbance-marginalized model can be regarded as the base distribution and then efficiently converted to the disturbance-aware model by invertible transformations $g$ conditioned on the disturbance.
\begin{align}
    p_w(s^\prime, r \mid s, a; d, \theta)
    &= p_w(g^{-1}(s^\prime, r; d) \mid s, a; \theta)
    \nonumber \\
    &\times \left | \cfrac{\partial g^{-1}(s^\prime, r; d)}{\partial s^\prime, r} \right |
\end{align}
With this design, even in the case of $\lambda \simeq 0$, the disturbance-marginalized model should be optimized through learning the disturbance-aware model.

However, the excellent representation capability of CNF is more than sufficient to compensate for prediction errors in the base distribution.
As a result, the two models might cause very different predictions.
Ideally, this is inappropriate because the prediction bias should not be induced by the disturbance, while the prediction variance can be increased.
It is possible to add a regularization term such that the two models match.
However, such a soft constraint would be difficult to adjust and the degree of divergence could be oscillatory, which is undesirable since LiRA works on the basis of the difference in predictive performance between the two.

Therefore, in LiRA, the above requirement (i.e. no prediction bias) is explicitly incorporated into the model design.
That is, the mean of the disturbance-marginalized model is carried over to that of the disturbance-aware model.
RNF proposed in the literature~\citep{kobayashi2023design} (see Fig.~\ref{fig:rnf}) makes it possible with the following restrictions:
\begin{itemize}
    \item The base distribution (in this case, the disturbance-marginalized model) is \textit{symmetric}.
    \item The invertivle transformations $g$ are \textit{odd} functions.
\end{itemize}
Thanks to RNF, while the two models produce generally similar predictions, their likelihoods differ depending on the presence or absence of disturbance:
namely, the larger the disturbance, the larger the variance of the disturbance-marginalized model, and the larger the gap.
As a result, the prediction gap between the two models should be stable and vary according to the disturbance intensity mainly, making it easier for the appropriate auto-tuning of $\lambda$ in LiRA.
In addition, since the probability distribution after transformation by RNF maintains a symmetric shape, the gradient to the disturbance that seeks to minimize its likelihood is likely to have an expected value of zero, preventing the excessively biased adversarial behaviors.

\subsection{Hindsight reparameterization gradient (HRG)}

\begin{figure}[tb]
    \centering
    \includegraphics[keepaspectratio=true,width=0.96\linewidth]{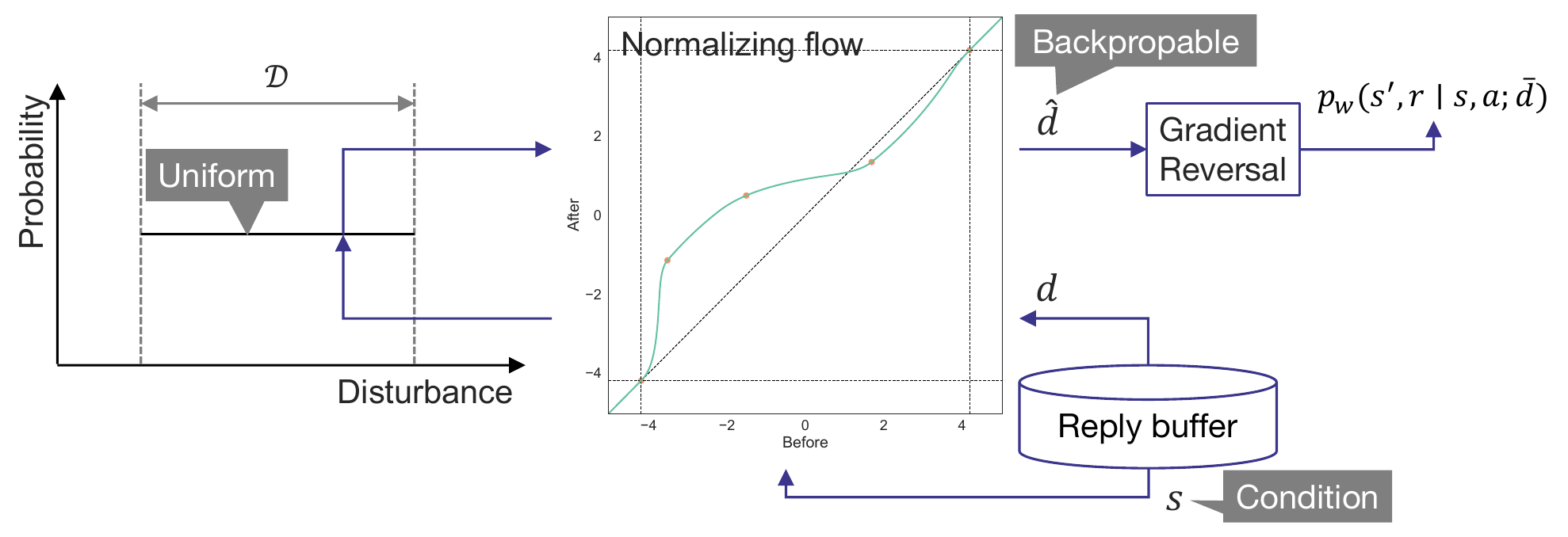}
    \caption{Hindsight reparameterization gradient (HRG):
    even if $d$ in the replay buffer has no computational graph, it can be obtained by passing through normalizing flows back and forth.
    }
    \label{fig:hrg}
\end{figure}

The adversary optimization is a challenge when solving the min-max problem of eq.~\eqref{eq:model_lira}.
That is, since it is indirectly involved in the prediction performance of models, the likelihood-ratio gradient (well-known as REINFORCE~\citep{williams1992simple}) is forced to be employed~\citep{mohaghegh2020advflow}.
However, this approach is known to be inefficient and sometimes unstable.

As shown in the literature~\citep{kingma2014auto,haarnoja2018soft}, a probability distribution can be efficiently optimized by a direct gradient of values sampled from it (in this paper, the disturbance $d$).
Such a so-called reparameterization gradient needs the computational graph in $d$ to $\phi$ (the parameters for adversary), and by applying GRL~\citep{ganin2016domain}, the adversarial learning defined as the corresponding min-max problem can be converted to the minimization problem simply.
However, in LiRA, this reparameterization gradient is only available by storing $d$ in a buffer while retaining its computational graph, or by learning in a streaming manner without the buffer.
The former wastes a lot of memory, and the latter significantly degrades efficiency.
In addition, it should be noticed that generating $d$ at each learning time without storing it in the buffer is inappropriate in LiRA because $s^\prime$ and $r$ are dependent on the disturbance at the time they experienced, which varies from the newly generated one.

To address this issue in implementing LiRA, a new trick, HRG, is proposed in this paper, as illustrated in Fig.~\ref{fig:hrg}.
Specifically, by focusing on the fact that the disturbance is box-constrained as $d \in \mathcal{D} = [-d^\mathrm{max}, d^\mathrm{max}]$, the probability distribution model for the adversary is designed with CNF, $g(d; s \phi)$, and a uniform distribution as its base $\mathcal{U}(-d^\mathrm{max}, d^\mathrm{max})$.
Note that CNF is a popular choice for adversarial learning due to its high expressive capability~\citep{mohaghegh2020advflow}.
Anyway, $d$ sampled from this adversary is stored in the buffer as a number with no computational graph.
When learning, if $\hat{d}$, which is numerically equivalent to $d$, can be resampled from the adversary with $\phi$ different from the time $d$ sampled, $\hat{d}$ can be utilized for learning as it is since it is with the computational graph to $\phi$ and equivalent to $d$ numerically.

To resample $\hat{d}$ for the given $d$, the following invertible transformations are performed.
\begin{align}
    \hat{d} = g(g^{-1}(d; s, \phi); s, \phi)
    \label{eq:hrg}
\end{align}
In that time, since $d$ is bounded, $\varpi(\hat{d}; s, \phi) \simeq 0$, which diverges the gradients w.r.t. $\phi$, can be naturally prevented.
In addition, by making the base distribution uniform, the gradients related to it are always zero, simplifying the computation.

Finally, the gradient of $\hat{d}$ is reversed by GRL as $\bar{d}$.
As a result, the loss function to be minimized for optimizing the disturbance-marginalized and -aware models are given as follows:
\begin{align}
    \ell = - \lambda \ln p_w(s^\prime, r \mid s, a; \theta) - (1 - \lambda) \ln p_w(s^\prime, r \mid s, a; \bar{d}, \theta)
\end{align}
Note that as mentioned when introducing Alg.~\ref{alg:lira}, the regularization term to the prior in eq.~\eqref{eq:model_lira} is calculated by Monte Carlo approximation using $\hat{d}$.

\subsection{Midrange-mean balancing (MMB)}

\begin{figure}[tb]
    \centering
    \includegraphics[keepaspectratio=true,width=0.96\linewidth]{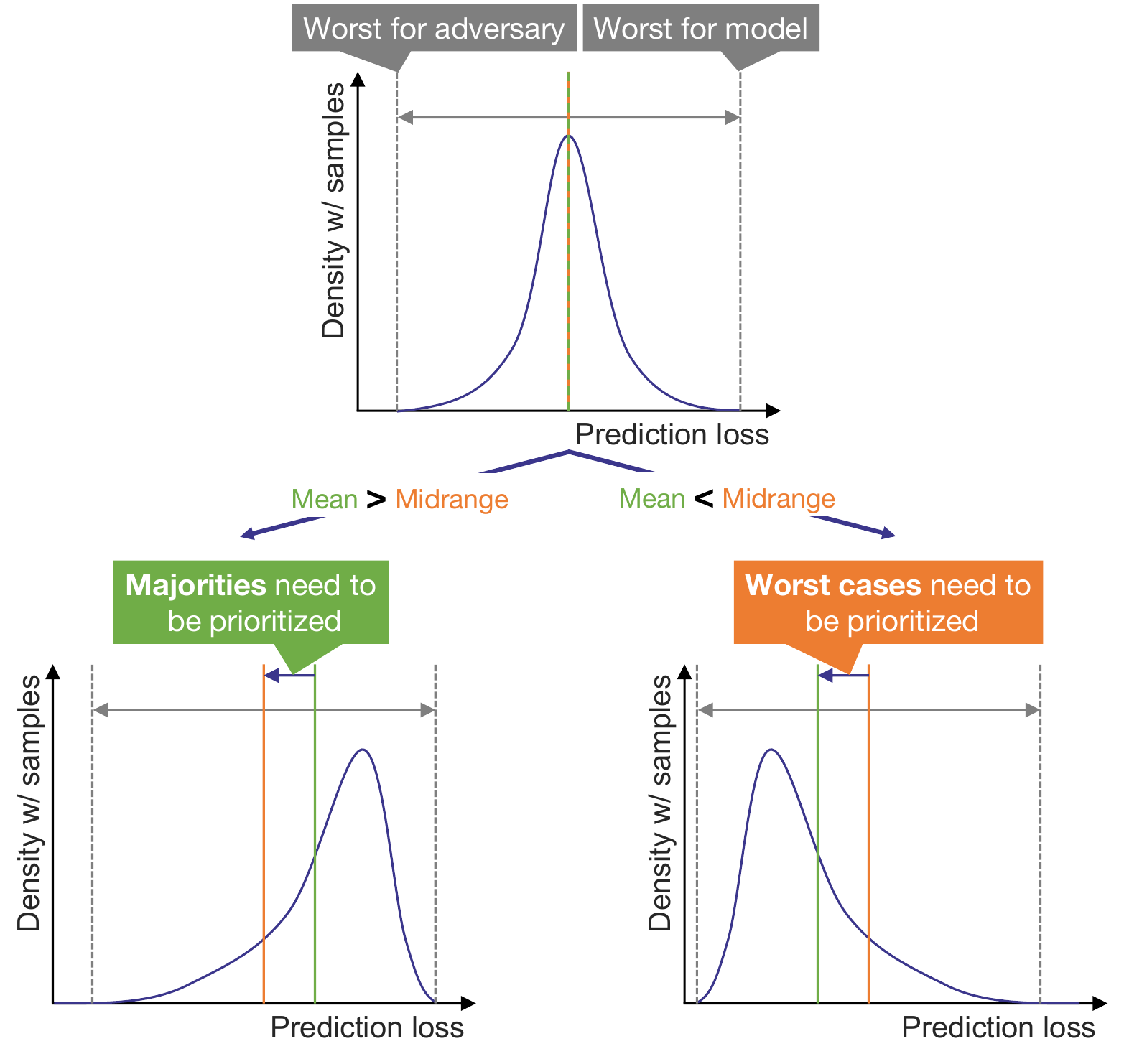}
    \caption{Midrange-mean balancing (MMB):
    the asymmetry in prediction loss can be captured, determining the prioritization to be minimized.
    }
    \label{fig:mmb}
\end{figure}

As mentioned above, to increase the robustness of the model (and MPC using it), rare events with low prediction accuracy should be prioritized to some extent~\citep{aotani2024cooperative}.
For adjusting the degree of prioritization, however, the previous work employed a meta-optimization method~\citep{aotani2021meta}, which requires an extra computational cost.
In addition, due to the adversarial learning nature of LiRA, it is expected that the low-quality data for the model and the adversary, which should be prioritized, might conflict with each other.

Therefore, this paper proposes a simpler automatic adjustment trick for the data prioritization, named MMB, keeping in mind that LiRA is a kind of adversarial learning.
Specifically, $N$ batch data have the corresponding losses $\{\ell_i\}_{i=1}^N$.
The standard way of minimizing them, i.e. the mean of losses that approximates the expected loss by Monte Carlo method, is first given as the main objective.
\begin{align}
    \mathcal{L}^\mathrm{mea} = \frac{1}{N} \sum_i \ell_i
\end{align}

In addition to this, MMB considers \textit{midrange}, which is a metric to represent central tendency like mean and median~\citep{dodge2003oxford}, as defined below.
\begin{align}
    \mathcal{L}^\mathrm{mid} = \frac{1}{2} (\min_{i} \ell_i + \max_{i} \ell_i)
\end{align}
That is, this midrange can simultaneously consider $\max_{i} \ell_i$, the worst case for the model, and $\min_{i} \ell_i$, the worst case for the adversary.

With this, the equality constraint, which aims to match the mean and midrange, is given.
\begin{align}
    \mathcal{L}^\mathrm{mea} = \mathcal{L}^\mathrm{mid}
\end{align}
This can be converted to the corresponding regularization term with an auto-tunable Lagrange multiplier $\gamma$, hence MMB minimizes the following losses.
\begin{align}
    \begin{split}
        \mathcal{L}^{\theta,\phi} &= \gamma \mathcal{L}^\mathrm{mid} + (1 - \gamma) \mathcal{L}^\mathrm{mea}
        \\
        \mathcal{L}^\gamma &= \gamma (\mathcal{L}^\mathrm{mea} - \mathcal{L}^\mathrm{mid})
    \end{split}
    \label{eq:mmb}
\end{align}
Here, $\gamma$ is generally in real space, but as in the case of $\lambda$, $\gamma \in [0, 1]$ is desired to avoid a direction that degrates the predictive performance of the models.
Note that the initial value of $\gamma$ is set to a small amout close to zero in order to efficiently improve the non-predictable model in the early stage of learning, so that the model behaves in a way that increases robustness after basic prediction performance is achieved.

Then, the qualitative behaviors of MMB are described below (also illustrated in Fig.~\ref{fig:mmb}).
\begin{itemize}
    \item $\mathcal{L}^\mathrm{mid} > \mathcal{L}^\mathrm{mea}$:
    The minimization of the mean drops some rare data, increasing the worst-case loss for the model.
    At the same time, there is a lot of data that should be disturbed by the adversary, reducing the prediction accuracy.
    As a result, MMB adjusts $\beta \to 1$ to emphasize the midrange, improving robustness.
    \item $\mathcal{L}^\mathrm{mid} < \mathcal{L}^\mathrm{mea}$:
    The model is poorly trained, with only a few data being highly predictive.
    In other words, majorities of data are difficult to predict with high accuracy (possibly due to excessive disturbances).
    As a result, MMB adjusts $\beta \to 0$ to emphasize the mean, making the prediction accuracy high on average.
    \item $\mathcal{L}^\mathrm{mid} \simeq \mathcal{L}^\mathrm{mea}$:
    It is well balanced with $\beta$ at that time.
\end{itemize}

\section{Numerical verification}

\subsection{Task}

\begin{figure}[tb]
    \centering
    \includegraphics[keepaspectratio=true,width=0.96\linewidth]{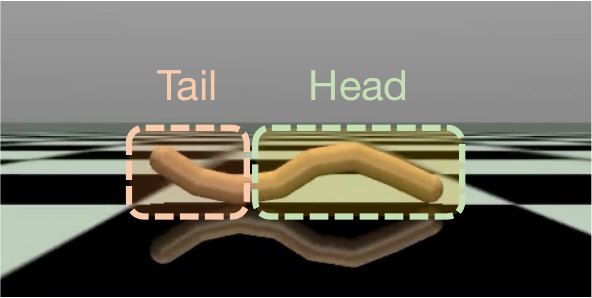}
    \caption{Worm-type robot simulated:
    4 joints on the head side are controllable, and the remaining 2--4 joints on the tail side are for generating disturbance.
    }
    \label{fig:robot_worm}
\end{figure}

Numerical simulations are conducted to verify the robot behaviors learned with the proposed LiRA.
Mujoco~\citep{todorov2012mujoco} is used as the simulator, and a task is to maximize the forward speed of a worm-type robot (see Fig.~\ref{fig:robot_worm}).
This robot locates 6--8 joints at equal intervals on its straight body: 4 controllable joints on the head side and 2--4 joints for disturbance on the tail side.
Note that the detailed task information with state, action, reward design are described in \ref{app:worm}.

The fewer the number of joints for disturbance, the smaller the effect of the disturbance and thus the less the need to limit the disturbance intensity;
in other words, the more the number of joints for disturbance, the more the task performance degradation and the more the need to limit the disturbance intensity.
That is, if the tolerance of the performance degradation is fixed for all the conditions, the disturbance intensities should converge to different values.

During learning, disturbances are generated by the agent's adversarial disturbance generator, but artificial disturbances must be applied to test the robustness of the learned policies.
Although a realistic disturbance could be a complex terrain surface, for reproducibility, artificial noises are added to the joints for the disturbances.
In this paper, the following three types of disturbances are prepared for testing.
Note that the prior in this task is given as Gaussian (a.k.a. nominal noise), $\mathcal{N}(0, \sigma_0=0.2/3)$.
The first is the nominal noise and is the weakest.
It is confirmed that such Gaussian noise tends to cancel each other out in the time evolution even when the scale is increased, and it is difficult to become the disturbances that significantly change the tail's posture.
Therefore, the second and third composes several Brownian noises $\epsilon^\mathrm{Brown}$, which have strong low-frequency components effective in changing the tail's posture, as below.
\begin{align}
    \begin{split}
        \tilde{d}_t &= \sum_{k=1}^K \frac{e^{\epsilon^\mathrm{Brown}_{k+K}}}{\sum_{\tilde{k}=1}^K e^{\epsilon^\mathrm{Brown}_{\tilde{k}+K}}} \epsilon^\mathrm{Brown}_k
        \\
        d_t &= k \frac{\sigma_0}{\sigma(\tilde{d})} \tilde{d}_t
    \end{split}
\end{align}
where $K=2$ is the number of noises mixtured and $k=3$ for the second and $k=6$ for the third denote the scale factors from the nominal noise.
$\sigma(\cdot)$ calculates the sample standard deviation along to pre-generated $\tilde{d}_t$ ($t=1,\ldots,T_{\mathrm{max}}$ with $T_{\mathrm{max}}$ the maximum time step for episode), not along to dimension of $d$.
Under the above disturbances, each model is tested 30 trials, and the interquartile mean of returns on the respective trials (referring to the literature~\citep{agarwal2021deep}) is used as a metric for each model.
In total, 12 models are learned under the same conditions with the respective random seeds, and then their noise-robust statistics (specifically, the best and worst cases are excluded at first, the outliers detected by the 1.5 IQR rule are also excluded\footnote{Due to the limited number of models, the interquartile mean, which excludes half of the data, was not employed.}, the remaining results are averaged with 95~\% confidence interval) is finally used as a metric for that conditions.

The specific learning conditions are in \ref{app:learn}.
During learning/inference, the robot is controlled with AccelMPPI~\citep{kobayashi2022real} (its conditions are in \ref{app:mpc}), but exploration noises are added to actions only during learning.
Note that $\lambda$, which corresponds to the adversary level, is give to be a scalar in the following simulations in order to focus on its tuning process and relationship between it and the sensitivity to disturbances, as mentioned above.

\subsection{Results}

\begin{figure*}[tb]
    \centering
    \includegraphics[keepaspectratio=true,width=0.96\linewidth]{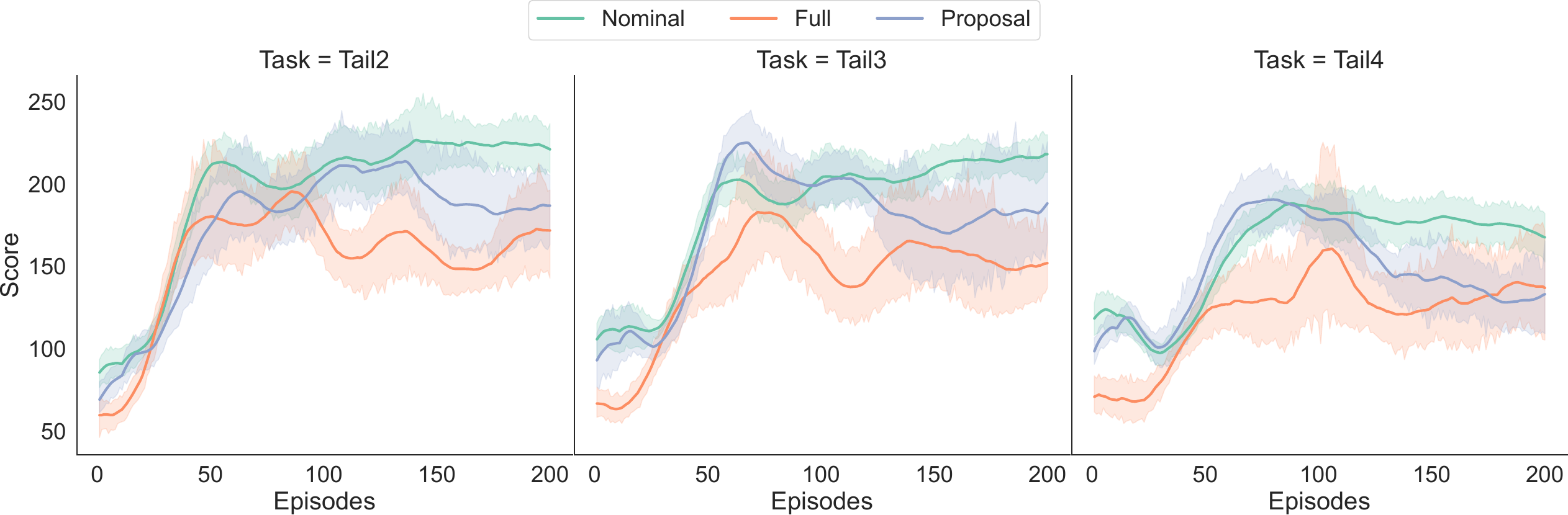}
    \caption{Learning curves of score (i.e. return):
    the implemented RL algorithm and adversarial learning work as expected.
    }
    \label{fig:compare_learn_score}
\end{figure*}

\begin{figure*}[tb]
    \centering
    \includegraphics[keepaspectratio=true,width=0.96\linewidth]{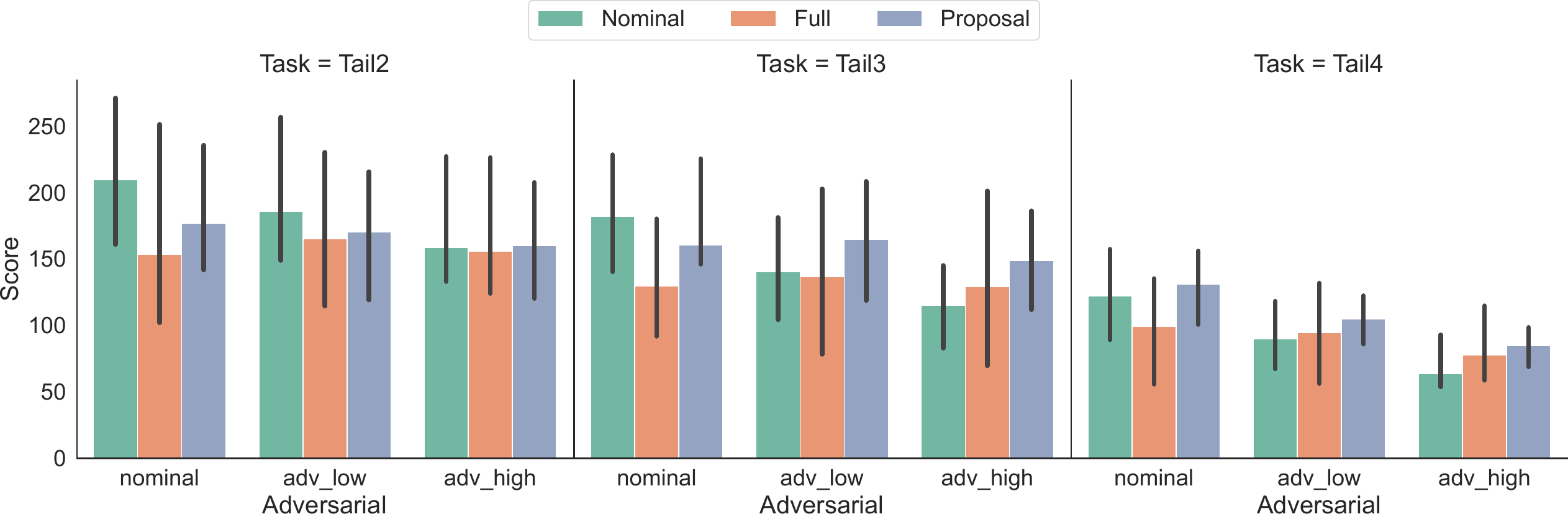}
    \caption{Task performance on three types of disturbance:
    \textit{Nominal} could not keep its task performance when the disturbance became large;
    \textit{Full} could keep its task performance while it was too conservative even with nominal noise;
    \textit{Proposal} obtained moderately robust and less conservative performance.
    }
    \label{fig:compare_summary}
\end{figure*}

\begin{table}[tb]
    \caption{Statistical performance evaluation by eq.~\eqref{eq:metric}}
    \label{tab:performance}
    \centering
    \begin{tabular}{l|ccc}
        \hline\hline
        Method & Tail2 & Tail3 & Tail4
        \\
        \hline
        \textit{Nominal} & -1044$\pm$349 & -768$\pm$265 & -556$\pm$167
        \\
        \textit{Full} & -994$\pm$391 & -759$\pm$327 & -394$\pm$197
        \\
        \textit{Proposal} & \textbf{-689$\pm$276} & \textbf{-553$\pm$232} & \textbf{-360$\pm$117}
        \\
        \hline\hline
    \end{tabular}
\end{table}

First, the following three conditions are compared in terms of the final task performance for the three types of disturbance.
\begin{itemize}
    \item \textit{Nominal} ($\lambda=1$, $\beta=\infty$):
    The model is learned only with the prior.
    \item \textit{Full} ($\lambda=0$, $\beta=0$):
    The model is learned in a fully adversarial manner.
    \item \textit{Proposal} ($\lambda$ is optimized, $\beta=10^{-3}$):
    The model is learned with LiRA.
\end{itemize}
Note that at the top priority in this paper is to confirm that the theoretically-derived LiRA behaves properly, comparisons are limited to that special cases and various existing methods are omitted.

First, to confirm that the RL algorithm in this study works properly, learning curves for the score are decpited in Fig.~\ref{fig:compare_learn_score}.
The low score at the beginning of the learning process increased after a certain number of episodes, indicating that the policy optimization by RL worked as expected.
The fact that the increase was low for \textit{Full} and decreases after the increase for \textit{Proposal} was due to the adversarial attacks.

Then, the test results are shown in Fig.~\ref{fig:compare_summary}.
It can be found that \textit{Nominal} without adversarial learning achieved high performance with small disturbance, but it was vulnerable to large ones.
\textit{Full}, which always learned adversarially to the maximum extent possible, maintained its performance independent of the disturbance intensity, but its basic performance was low and conservative.
Compared to \textit{Nominal}, \textit{Proposal} (a.k.a. LiRA) was able to suppress the performance degradation caused by the disturbance intensity, was less conservative than \textit{Full}, and succeeded in achieving a well-balanced and stable learning.

The following metric is defined to quantitatively evaluate the conservativeness reduction and the robustness improvement.
\begin{align}
    &(S^\mathrm{nominal} - \max_\mathrm{methods,models} S^\mathrm{nominal})
    \nonumber \\
    +& \sum_\mathrm{i \in noises} (S^\mathrm{i} - \max_\mathrm{models} S^\mathrm{nominal})
    \label{eq:metric}
\end{align}
where, $S$ denotes the return.
The results of this statistical evaluation are summarized in Table~\ref{tab:performance}.
\textit{Nominal} and \textit{Full} did not have high values in any of the tasks due to low robustness or excessive conservativeness.
In contrast, \textit{Proposal} had the highest values in all tasks, and in addition, its standard deviations were smaller than the others.

\begin{figure*}[tb]
    \centering
    \includegraphics[keepaspectratio=true,width=0.96\linewidth]{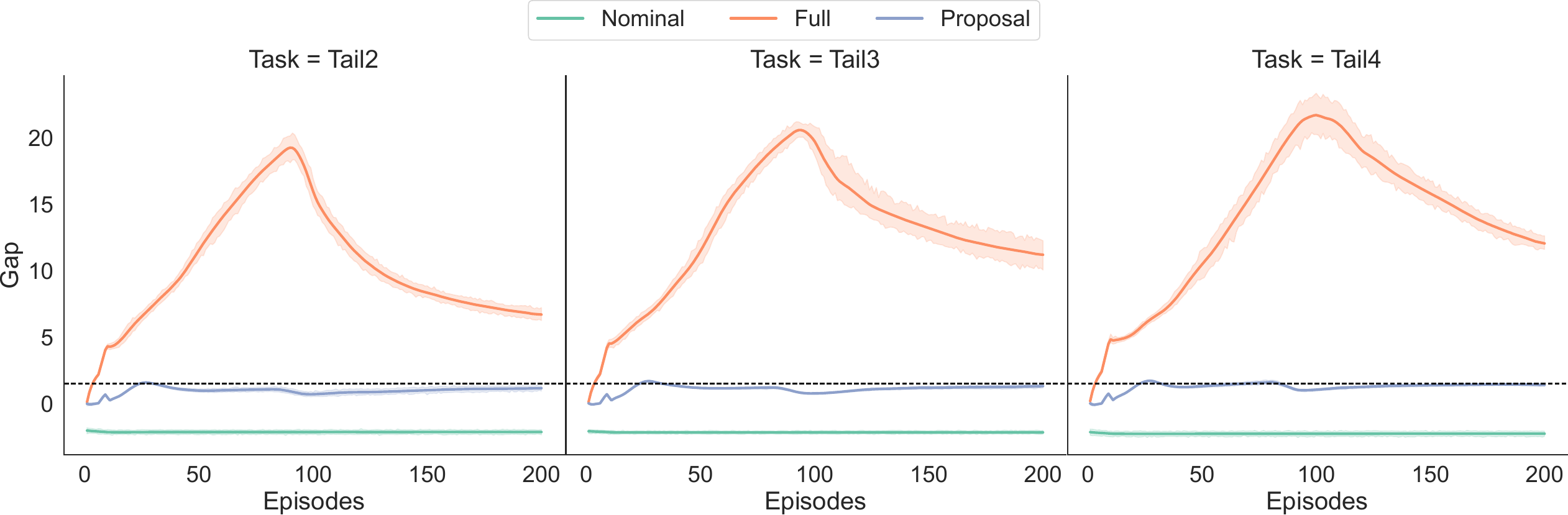}
    \caption{Learning curves of $\ln p_w(s^\prime, r \mid s, a; \theta) - \ln p_w(s^\prime, r \mid s, a; d, \theta)$:
    the proposed method appropriately satisfied the light-robust constraint.
    }
    \label{fig:compare_learn_gap}
\end{figure*}

Next, the process, by which this appropriate balance was achieved, is analyzed.
To do so, the gap of prediction performances with/without disturbance information (corresponding to the constraint for the light robustness), $\ln p_w(s^\prime, r \mid s, a; d, \theta) - \ln p_w(s^\prime, r \mid s, a; \theta)$, is first shown in Fig.~\ref{fig:compare_learn_gap}.
In \textit{Nominal}, $\ln p_w(s^\prime, r \mid s, a; \theta) > \ln p_w(s^\prime, r \mid s, a; d, \theta)$ was caused since RNF was not optimized at all, adding a bias to the prediction.
In \textit{Full}, the gap was too large due to the excessive disturbance, which significantly reduced the prediction performance without disturbance information.
In contrast, \textit{Proposal} generally converged on the pre-specified threshold, $\rho$, for all tasks.
This indicates that the light robustness constraint is properly satisfied by the proposed implementation.
Note that the gap did not always equivalent to $\rho$ because the light-robust constraint is given as an inequality constraint (in addition, a limitation of numerical optimization).

\begin{figure*}[tb]
    \centering
    \includegraphics[keepaspectratio=true,width=0.96\linewidth]{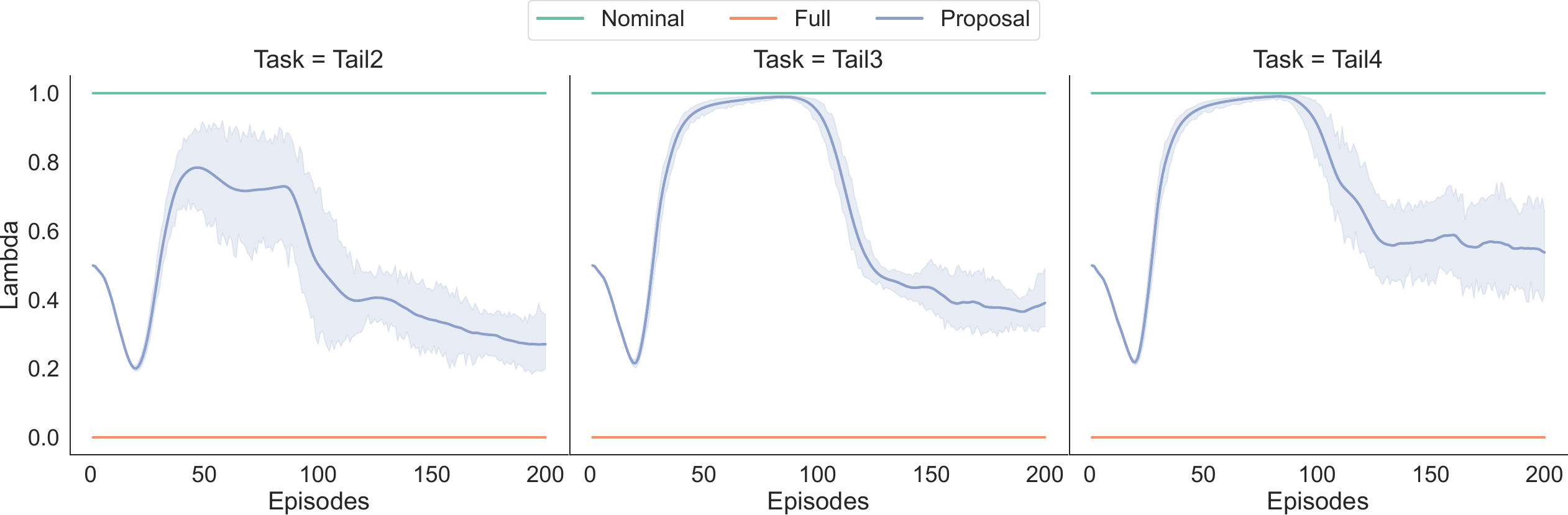}
    \caption{Learning curves of $\lambda$ auto-tuned by LiRA:
    a curriculum was naturally formed that weakened the disturbance until the middle of the learning process and strengthened it later, and the final adversary level was greater as the effect of the disturbance was weaker.
    }
    \label{fig:compare_learn_lambda}
\end{figure*}

In order to satisfy this constraint, how $\lambda$ was automatically tuned by LiRA is depicted in Fig.~\ref{fig:compare_learn_lambda} with its statistics summarized in Table~\ref{tab:balance}.
Note again that the smaller $\lambda$ is, the stronger the disturbance intensity.
The overall trend is that the fewer the number of disturbance joints and the smaller the original disturbance influence, the smaller $\lambda$ was, giving priority to adversarial learning.
Conversely, as the number of joints for external disturbances increased, $\lambda$ became larger, suggesting that the disturbance intensity was suppressed to prevent the performance degradation.

Looking at the early training period, we see that $\lambda$ was updated toward zero once in every trial.
This is due to the fact that the initialized adversary was not subjected to a significant collapse of the models' prediction accuracy, and the the disturbance-aware model was also not able to properly reflect the effects of disturbances in its predictions.
After getting proper functions of them, $\lambda$ was once large from the beginning to the middle of learning progress, and after a while, it became smaller and converged to each value suitable for the task.
This result suggests the emergence of some kind of curriculum that once simplifies the task to easily predictable situations and then makes the task more difficult by gradually increasing the disturbance.
Indeed, it can be said that the derivation of LiRA is partially similar to the self-paced learning methods~\citep{kumar2010self,klink2020self}, and therefore, LiRA might lead to such an additional value.
That is, the disturbance-aware model in LiRA can be considered as an automatic pacemaker for such a curriculum.
As another characteristic in the learning process, it can be found from the confidence intervals that when $\lambda$ started to converge statistically in the second half of the learning, the fluctuation of $\lambda$ was active, implying that perhaps adversarial learning reached an unstable equilibrium.

\begin{table}[tb]
    \caption{Statistics of $\lambda$ auto-tuned by LiRA}
    \label{tab:balance}
    \centering
    \begin{tabular}{ccc}
        \hline\hline
        Tail2 & Tail3 & Tail4
        \\
        \hline
        0.474$\pm$0.279 & 0.630$\pm$0.282 & 0.694$\pm$0.274
        \\
        \hline\hline
    \end{tabular}
\end{table}

\subsection{Ablation tests}

\begin{figure}[tb]
    \centering
    \includegraphics[keepaspectratio=true,width=0.96\linewidth]{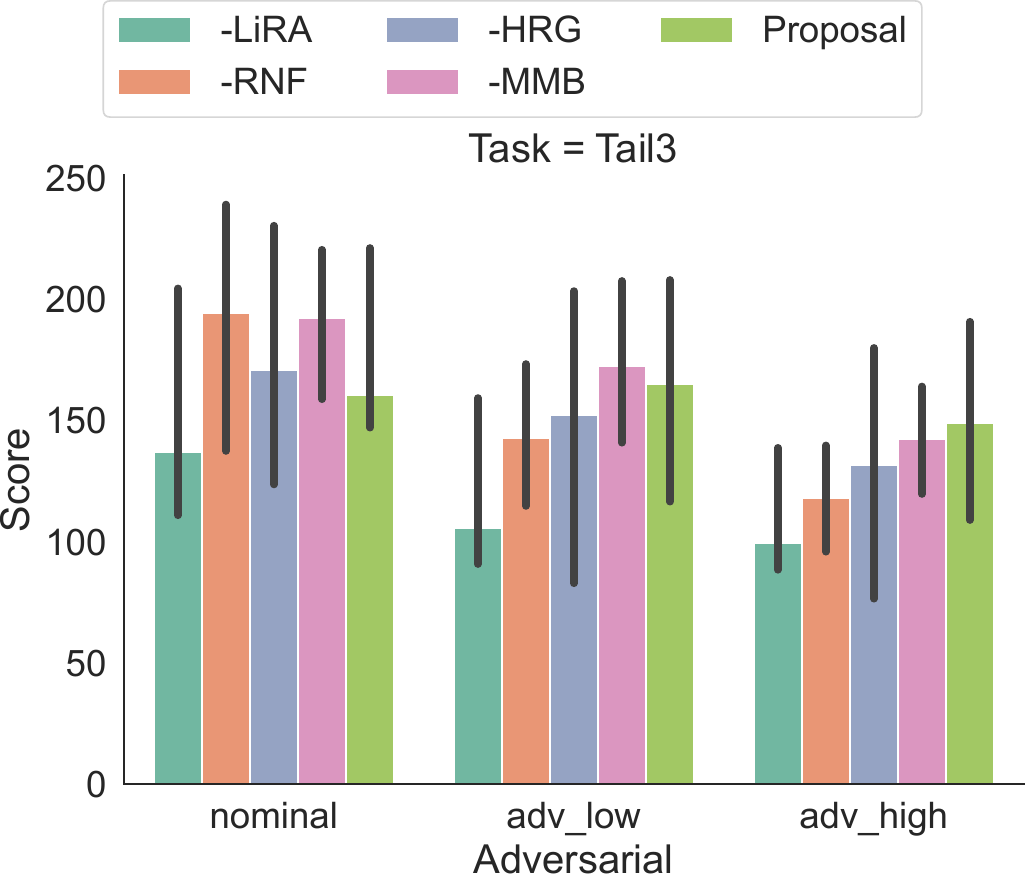}
    \caption{Task performance on ablation tests:
    only \textit{-MMB} and \textit{Proposal} achieved the good scores on average.
    }
    \label{fig:ablation_summary}
\end{figure}

\begin{figure}[tb]
    \centering
    \includegraphics[keepaspectratio=true,width=0.96\linewidth]{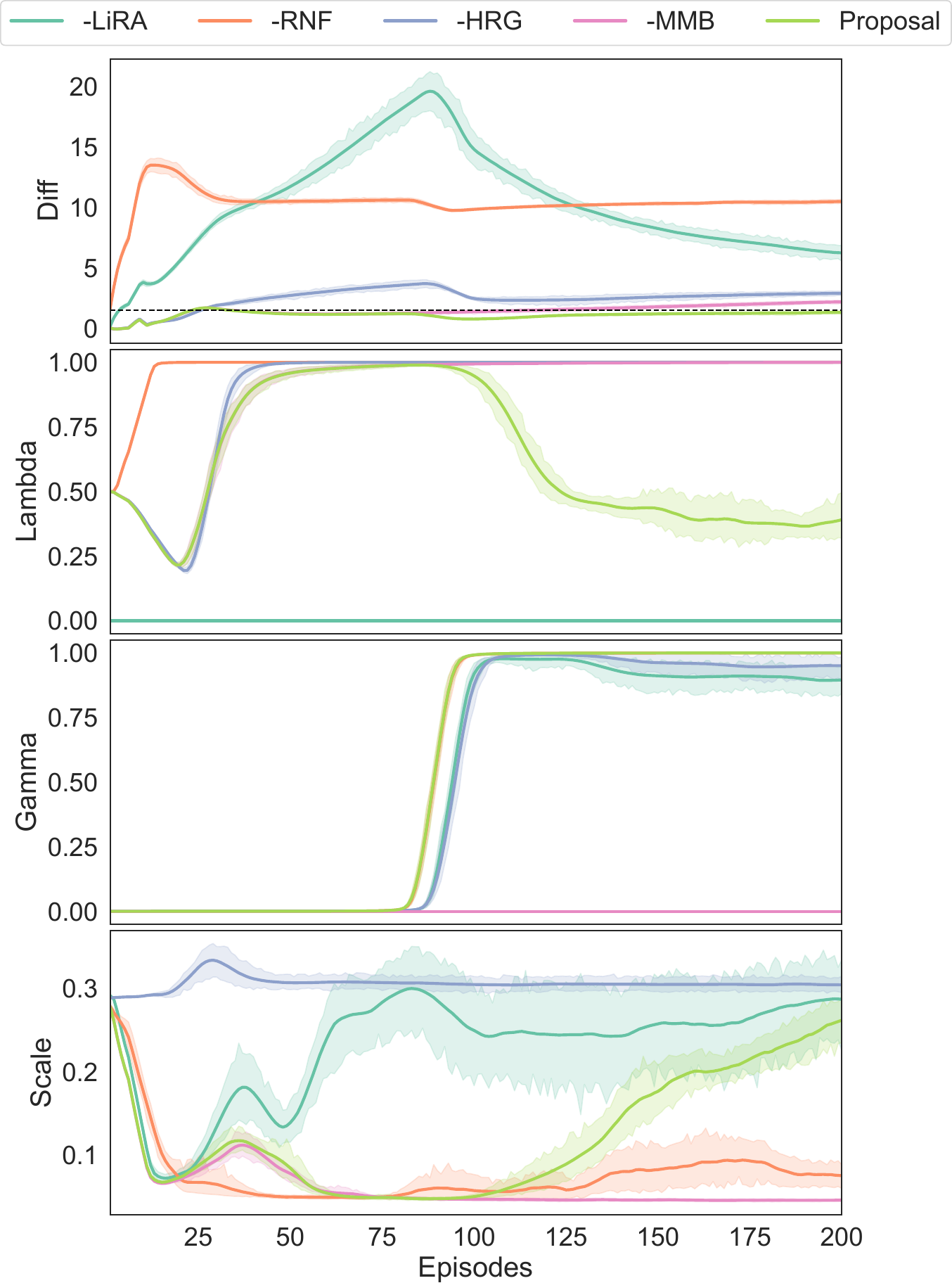}
    \caption{Learning curves on ablation tests:
    \textit{-MMB} failed to auto-tune the adversary level, while \textit{Proposal} succeeded.
    }
    \label{fig:ablation_learn}
\end{figure}

To gain a deeper understanding of LiRA, ablation tests, which partially exclude implemented functions, are conducted.
Specifically, the following conditions are compared to \textit{Proposal} in the above section:
\begin{itemize}
    \item \textit{-LiRA}:
    The light-robust constraint is omitted.
    \item \textit{-RNF}:
    Instead of RNF, the standard CNF is applied.
    \item \textit{-HRG}:
    Instead of HRG, the likelihood-ratio gradient is applied.
    \item \textit{-MMB}:
    Only the mean of losses is minimized.
\end{itemize}
For simplicity, only the Tail3 task, which is medium affected by disturbances, is learned here.

The test results are shown in Fig.~\ref{fig:ablation_summary} and the learning behaviors are depicted in Fig.~\ref{fig:ablation_learn}.
Note that the upper graph illustrates the same metric as that of Fig.~\ref{fig:compare_learn_gap} and \textit{Scale} in the bottom graph means the sample standard deviation of $d$.
\textit{-LiRA} was with the worst performance since it is equivalent to \textit{Full} except existence of the regularization to the prior, namely, \textit{-LiRA} made the policy too conservative.

Except the case with the nominal noise, \textit{-RNF} and \textit{-HRG} also degraded the performance compared to \textit{Proposal}.
This can be explained with their learning progress.
That is, \textit{-RNF} violated the light-robust constraint, that is the gap could not be reduced even with $\lambda \to 1$.
In addition, \textit{-HRG} failed to optimize the adversary, and the scale could not converge to around the prior scale even with $\lambda \to 1$.
As a result, they failed to improve the policy robustness without proper adversarial learning.
Hence, these tricks prove to be indispensable for LiRA.

In contrast, \textit{-MMB} outpeformed \textit{Proposal} when the disturbance intensity was not strong, while \textit{Proposal} was more robust than \textit{-MMB}, as can be seen in the strongest disturbance intensity.
This is natural because MMB prioritized the worst-case data, making the model (and the corresponding policy accordingly) robust.
In addition, \textit{-MMB} converged to $\lambda \to 1$ eventually, not experiencing strong disturbances.
Indeed, when MMB started to prioritize the worst-case data around 100th episode, the performance gap became narrow once, leading to an opportunity to reduce $\lambda$.
Thus, it is indicated that MMB is also critical for acquiring the desired moderate robustness.
Note that $\gamma$ converged to $1$ in this simulation, but this is not always the case.
For example, in the following real-robot demonstration, the converted value was around 0.9.

\section{Real-robot demonstration}

\subsection{Task}

\begin{figure}[tb]
    \centering
    \includegraphics[keepaspectratio=true,width=0.96\linewidth]{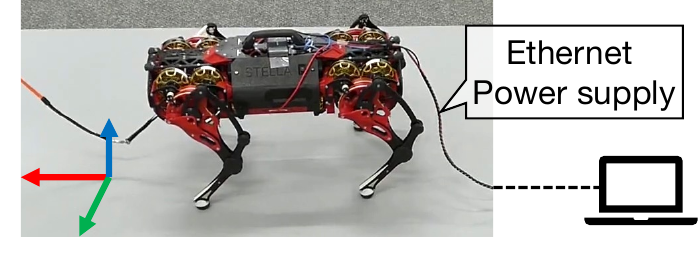}
    \caption{Quadrupedal robot for demonstration:
    as a base controller, it has a periodic walking motion generator and an impedance model;
    their stride and stiffness on the horizontal plane are robustly optimized according to disturbance.
    }
    \label{fig:robot_quad}
\end{figure}

LiRA is demonstrated in the real world with a quadrupedal robot (see Fig.~\ref{fig:robot_quad}).
The robot has an oscillator-based walking generator and an virtual impedance model for external forces as motion primitives, but their stride and stiffness should be optimized adaptively.
For example, if there is no disturbance, it would be good for the robot to stand in place;
if small disturbances are injected, increasing the stiffness might be able to resist them;
and walking should be activated for following large disturbances.
This demonstration confirms that such behaviors can be empirically acquired by LiRA.

The details of this task configuration and learning/MPC conditions are all described in \ref{app:quad}, \ref{app:learn}, and \ref{app:mpc}, respectively.
In summary, the adversary applies virtual horizontal forces to the robot body as like~\citep{shi2024rethinking}, and then, LiRA learns the model for optimizing the horizontal stride and stiffness.
The nominal-level disturbance sets so as to have little effect on the robot, not allowing it to learn how to respond to the disturbance at all, while the disturbance near its maximum intensity, which is adversarily given, can easily make the robot topple, causing a high risk of malfunction\footnote{Indeed, the excessive disturbance given during debugging caused the robot to tip over and damage the frame.}.
That is, both \textit{Nominal} and \textit{Full} are not suitable for this task.
In contrast, LiRA is expected to safely achieve the moderate robustness by controlling the disturbance intensity so as not to cause such undesired situations.

The performance of the model learned by LiRA is verified by applying two types of periodic virtual horizontal forces to ensure reproducibility.
The first disturbance is a nominal noise (i.e. $\sim$30~N), and the second is $\sim$150~N, which is almost equivalent to the robot's weight.
The change in the robot behaviors against these disturbances are tested as the learning progresses.
In addition, the behavior against the real-world disturbance given by a human is illustrated.

\subsection{Results}

\begin{figure}[tb]
    \centering
    \includegraphics[keepaspectratio=true,width=0.96\linewidth]{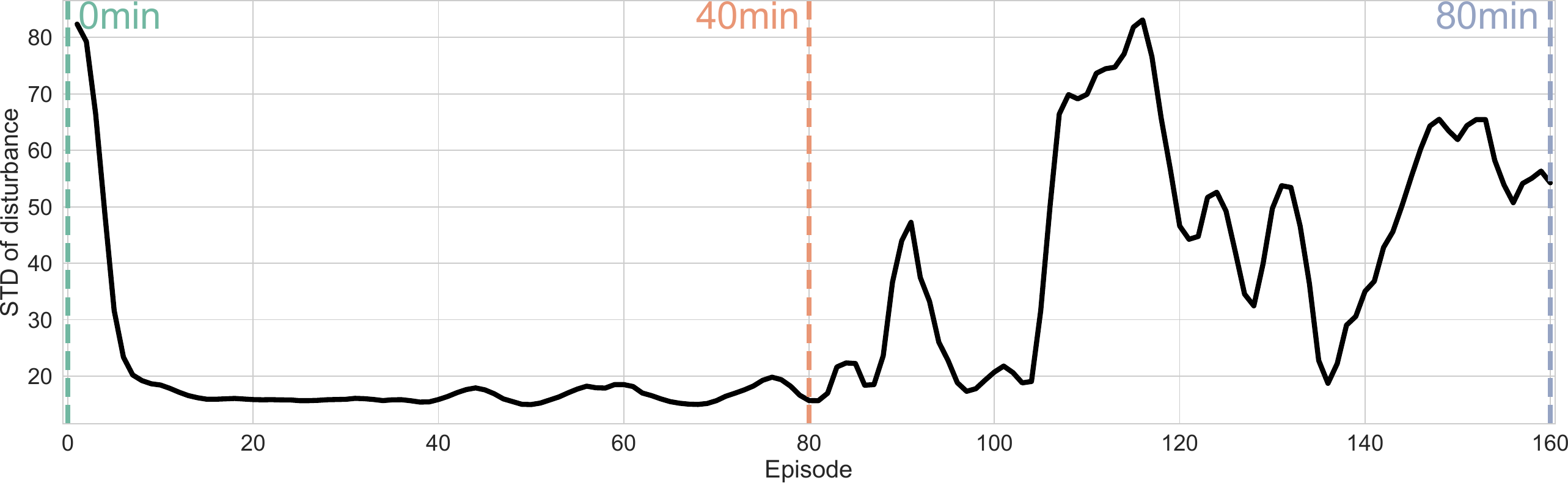}
    \caption{Standard deviation of disturbance at learning phase:
    in the first half of learning, as many unknown states are difficult to be predicted, the disturbance intensity was suppressed to almost nominal noise;
    in the second half of learning, the model with enough prediction accuracy allowed for increasing the disturbance intensity.
    }
    \label{fig:exp_learn}
\end{figure}

First, the scale (i.e. standard deviation) of the disturbance induced by the auto-tuned adversary level in LiRA is plotted in Fig.~\ref{fig:exp_learn}.
Note that one episode has a maximum of 600 steps, which correspond to 30 seconds of RL internal time, and a total of 160 episodes (approximately 80 minutes) were performed in learning.
Similar to the simulation results, LiRA suppressed the disturbance to the nominal noise level in the first half of learning, during which time the basic dynamics would be learned.
The disturbance intensity was then increased to allow the robot to experience a variety of disturbances.
Note that the disturbance intensity never changed monotonically in the second half of learning.
This may be due to the fact that $\lambda$, which suggests the adversary level, is defined as a state-dependent function in this experiment: namely, when the robot encountered unknown states, the adversary level for that was rapidly suppressed once.

The responses and performances to the periodic virtual external forces are investigated using the models at the following three characteristic times during this learning process.
\begin{itemize}
    \item \textit{0min}:
    The model at the beginning should has no knowledge how to respond the disturbance properly.
    \item \textit{40min}:
    The model at the middle was trained almost only on the nominal noise like \textit{Nominal}.
    \item \textit{80min}:
    The model at the end was trained properly on the various disturbance intensities and directions.
\end{itemize}

\begin{figure*}[tb]
    \centering
    \includegraphics[keepaspectratio=true,width=0.96\linewidth]{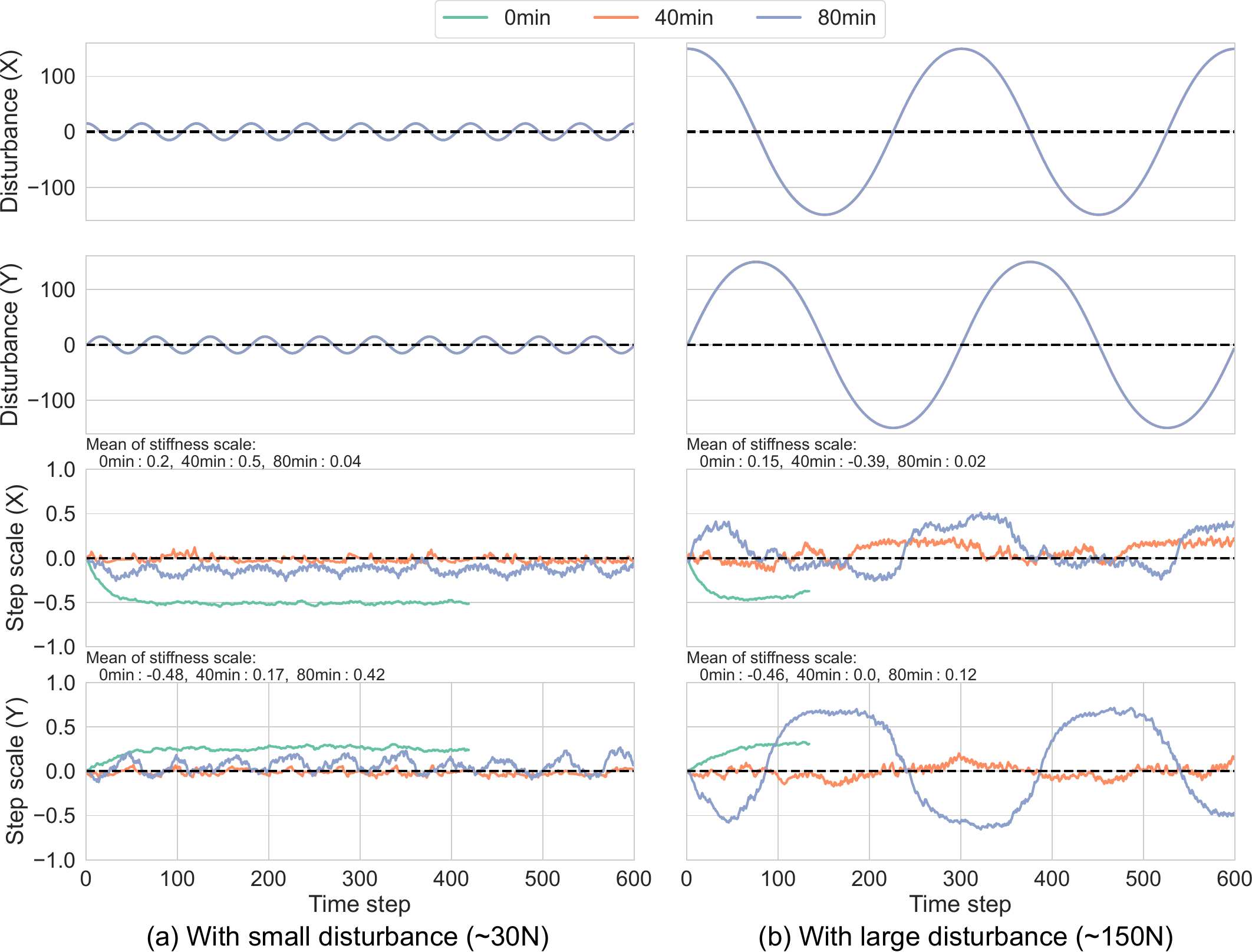}
    \caption{Tests with two types of periodic disturbance:
    with the unlearned model (i.e. \textit{0min}), the robot is operated independently of the virtual external forces applied, failing to continue the tests;
    \textit{40min}, which only learned up to its response to the nominal noise, mainly selected to stand in place regardless of the disturbance intensity, although it decreased the stiffness for the large disturbance;
    the model after learning \textit{80min} could stand almost in place for the small disturbance, while it walked to follow the large disturbance with a slight phase delay.
    }
    \label{fig:exp_disturb}
\end{figure*}

The results of adding the two types of disturbances described above to these models are depicted in Fig.~\ref{fig:exp_disturb} (also see in the attached video).
Naturally, the model at \textit{0min} determined the wrong strides (and stiffnesses) regardless of the disturbance, and as a result, the tests were terminated when it went out of the field or was judged to tip over.
The strategy in training \textit{40min} mostly with the nominal noise was optimized for the small disturbances, that is, it increased the stiffness to some extent to resist the disturbance and basically stood in place.
On the other hand, when given the large disturbances, which were rarely experienced by that time, the robot's body was made to move in accordance with the disturbance by reducing the stiffness, but the robot still tried to stand  in place, losing its balance.

With the model at \textit{80min}, the similar behavior for the small disturbance to the model at \textit{40min} was observed.
However, the difference is that the robot was prepared for some large disturbances just in case by taking slight steps backward, perhaps because the asymmetrical structure of the robot makes it prone to backward imbalance.
This difference in behavior yielded the safe strategy for the large disturbance: namely, the stiffness was suppressed to make it easier to capture the effect of the disturbance, and the robot walked in the direction of the disturbance to follow it.
As a result, the return was increased from 682 in \textit{40min} to 703 in \textit{80min}.

\begin{figure}[tb]
    \centering
    \includegraphics[keepaspectratio=true,width=0.96\linewidth]{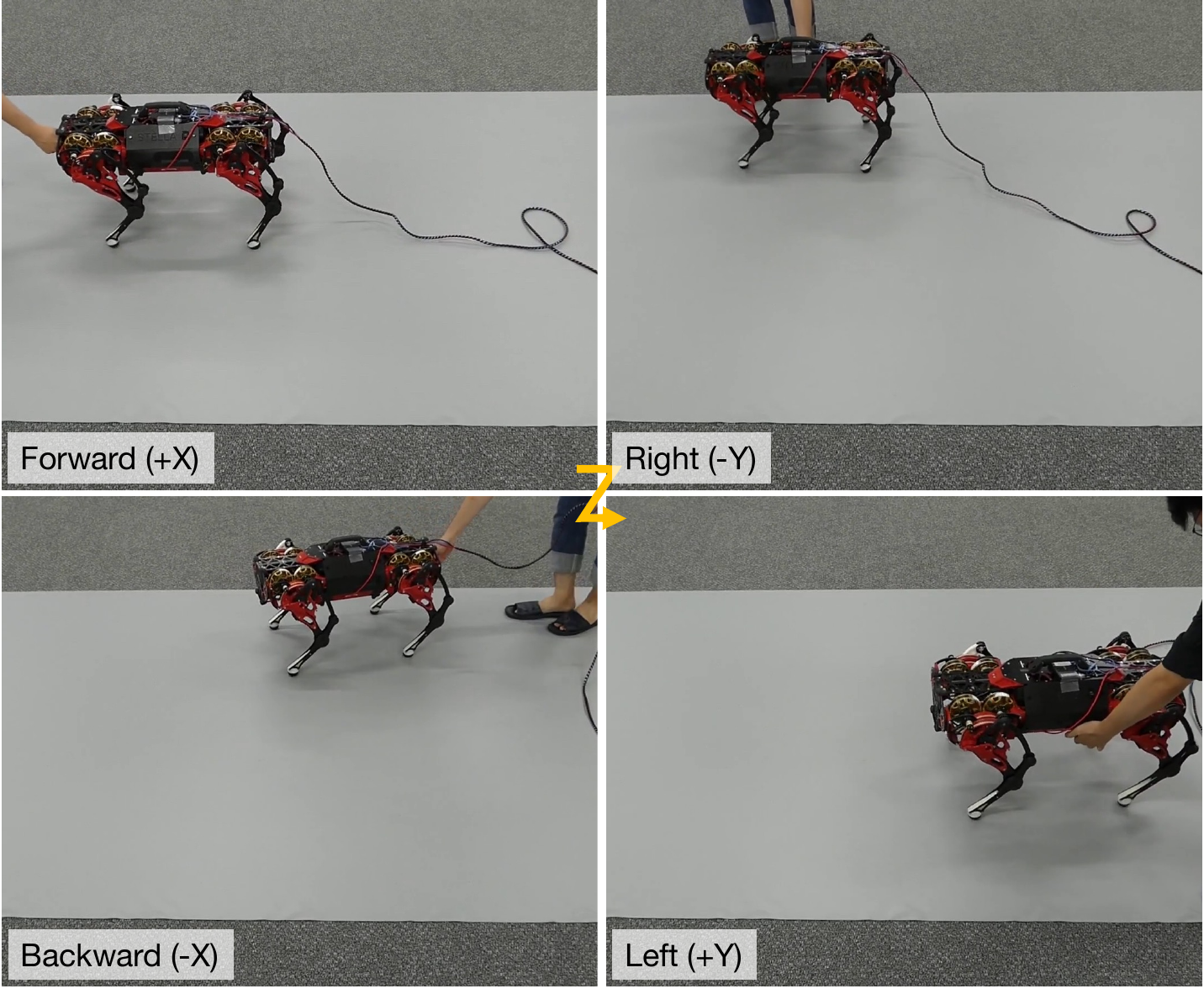}
    \caption{Snapshots of demonstration with real-world disturbance:
    the model learned by LiRA successfully mitigated the effects of the disturbance given by human pulling as it walked in the direction of the disturbance.
    }
    \label{fig:exp_snap}
\end{figure}

Finally, the model learned by LiRA is tested whether it can acquire adaptive behavior under the real-world disturbances.
That is, as shown in Fig.~\ref{fig:exp_snap}, the robot is pulled on its body in four directions by a human.
As can be found in the attached video\footnote{also see here \url{https://youtu.be/63PDhDfWKf0}}, the robot successfully walked in the directions of the disturbances and maintained its balance as expected from the results at the case with the large virtual disturbances.

\section{Discussion}
\label{sec:discussion}

\subsection{State dependency of adversary level}

\begin{figure*}[tb]
    \centering
    \includegraphics[keepaspectratio=true,width=0.96\linewidth]{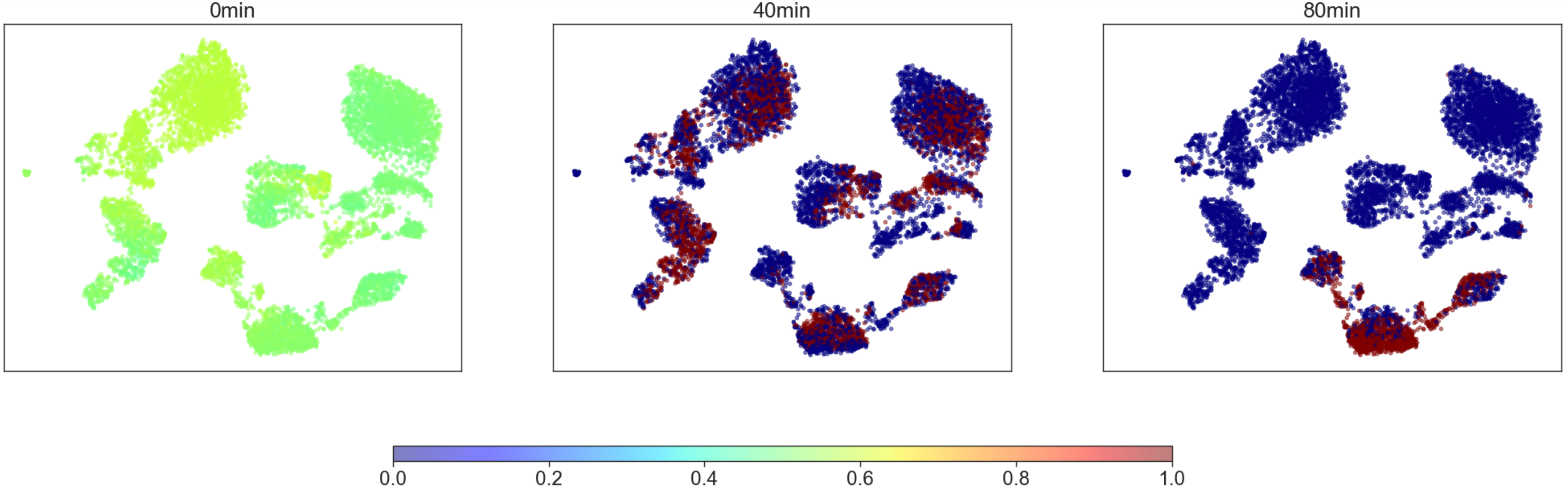}
    \caption{$\lambda$ for states experienced (embedded into two dimensions by UMAP~\citep{mcinnes2018umap}):
    $\lambda$ was initialized around $0.5$;
    as LiRA progressed, $\lambda$ was polarized to zero or one, and gradually $\lambda \simeq 0$ increased;
    the cluster of states with $\lambda \simeq 1$ was still left at the end of learning probably as the states losing balance, which are difficult to predict future states.
    }
    \label{fig:exp_lambda}
\end{figure*}

As mentioned above, the inverse of the adversary level, $\lambda$, can be learned as a state-dependent function, as in the real-world demonstration above.
In this case, not only the time variation of $\lambda$ but also the state-dependent differences should be confirmed.
Therefore, $\lambda$ at each learning progress is illustrated in Fig.~\ref{fig:exp_lambda} for the states in the buffer finally collected during the above demonstration, which are embedded into two dimensions by UMAP~\citep{mcinnes2018umap}.
At the start of learning, $\lambda$ was initialized to be near $0.5$ in all states.
$\lambda$ was polarized to zero or one as learning progressed, and $\lambda \simeq 0$ increased gradually.
In other words, the inexperienced states at \textit{40min} were with $\lambda \simeq 1$ at that time, and then, they were encountered by the enhanced disturbances, shifting $\lambda$ to zero by learning the way of predicting the future states from them.
However, even at the final stage \textit{80min}, there was still a cluster of states with $\lambda \simeq 1$.
This cluster probably represents states out of balance and cannot increase their adversary level because of low predictability.
As a result, disturbances that would intentionally tip over might be suppressed, making it easier to ensure safety.

Thus, the state-dependent $\lambda$ basically functions as expected, but some issues remain.
First, at \textit{40min}, the states with $\lambda \simeq 0$ and $\lambda \simeq 1$ were mixed, suggesting that a slight change in state might cause a sudden change in $\lambda$.
Since such a sudden change in $\lambda$ induces instability in learning, it is desirable to smooth the function of $\lambda$, for example, as in the literature~\citep{kobayashi2022l2c2}.
Furthermore, in the cluster that seems to be out of balance, some states with $\lambda \simeq 0$ at \textit{40min} were reverted to $\lambda \simeq 1$ at \textit{80min}.
Since this is considered to be a failure in the estimation of $\lambda$, it would be better to update $\lambda$ more carefully, for example, by considering uncertainty in parameter estimation based on Bayesian theory~\citep{wenzel2020hyperparameter}.
Finally, although the initial value of $\lambda$ was near $0.5$, it would be more natural to start with the larger $\lambda$.

\subsection{Practical and theoretical time efficiency}

It is noteworthy that even though the motion primitives were prepared in the above real-world experiment, the moderate robustness to disturbances was achieved in a short time of data collection (80 minutes).
Strictly speaking, however, this time is RL internal time spent on data collection, and the actual experiment time is different.
First, MPC was conducted to keep 20~fps as much as possible, but in reality there was a slight excess sometimes, and taking into account the subsequent communication and measurement delays, there was a difference of a little less than 1.5 times between the internal and actual times.
The learned model was able to predict state transitions and rewards without problems because it was trained under this difference, but the actual time required for data collection was just barely less than two hours.
It would be important to compensate this gap by implementing a strict real-time control system.

On the other hand, as summarized in Alg.~\ref{alg:lira}, the implementation of LiRA in this paper is divided into the data collection phase and the learning phase, so the actual experiment time should include the time of the learning phase.
After the end of each episode, all the data in the buffer is replayed once for learning, so the learning time inevitably increased as the number of data increased, eventually taking more than one minute per each.
Therefore, the actual total experiment time, including robot initialization (with cooling actuators), exceeded six hours.
Although it would be possible to use a multiprocessing for learning the model behind the data collection phase, this may affect control performance because the computational resources needed for MPC optimization must be allocated to learning.
In addition, it must be said that HRG in the implementation of LiRA incurs extra computational costs, so it may be necessary to review the implementation of LiRA to make it faster.

The above issues regarding time efficiency are mainly due to the practical implementation, but apart from that, further theoretical contributions can be considered.
First, the exploration by the RL agent in this implementation relies on Gaussian noise optimized by MPC, but it is known that such a random exploration is not efficient~\citep{wilson2021balancing}.
As suggested in the previous studies such as~\citep{seo2021state}, the exploration efficiency could be improved by actively selecting the actions that can transition to states with high uncertainty from the vicinity of its likely optimal actions.
The adversary may also be better to prioritize the disturbances that transition to states where its adversary level is uncertain in conjunction with making $\lambda$ stochastic suggested above.
By such, the adversary could reduce the unnecessary attacks.

\subsection{Slowdown of auto-tuning in overlearning}

In the real-robot experiment, LiRA succeeded in achieving the moderate robustness of the model by gradually increasing the disturbances.
At the end of 80 minutes (i.e. 160 episodes), $\lambda$ had converged, but as suggested by the simulations, this is expected to be on an unstable equilibrium point.
Therefore, it should be additionally checked whether this balance is lost when overlearning is performed, and whether that can be recovered by readjusting $\lambda$.
Finally, 120 minutes (i.e. 240 episodes) of learning was conducted.
As a result, the disturbances became more aggressive to make the model more unpredictable, causing the robot to lose its balance and fail episodes more frequently.
Such a situation would normally be avoided by readjusting $\lambda$ so that the disturbances are again suppressed to restore balance.
However, in this additional experiment, the increase in $\lambda$ became very slow and the disturbances could not be satisfactorily suppressed until the end.

The reason for this slowdown in the auto-tuning of $\lambda$ might be because the proportion of worst-case data in the buffer was smaller.
In other words, the learning of normal data, which is the majority, inhibited the reflection of the worst-case data.
Although MMB was expected to mitigate this situation, the worst cases were rarely included in small batches, and the worst cases could not be fully considered.

To address this limitation, a technique that more explicitly considers the value of the data is needed instead of (or in addition to) MMB.
For example, several techniques have been proposed to prioritize the replay of data from the replay buffer~\citep{oh2022model,pan2022understanding}, and inspired by them, it would be effective to prioritize the replay of data with poor model prediction accuracy (or data with insufficient interference by the adversary).
Alternatively, a weighting that is theoretically equivalent to the prioritization of replay data would also prevent the slowdown.
One solution may be a method that utilizes the softmax operator as the weighting~\citep{lin2024smooth}.

\subsection{Heuristic design}

In the theoretical derivation and practical implementation of LiRA, several hyperparameters need to be specified.
This paper set them empirically.

In particular, the tolerance of performance degradation between the disturbance-aware and -marginalized models, $\rho$, has empirical lower and upper bounds, and then it was searched within that range for obtaining a reasonable $\lambda$ in the Tail3 simulation.
In other words, the lower bound of $\rho$ is given to be the degradation when no disturbance is given, which is attributed to the additional expressive power of RNF.
The upper bound of $\rho$ is roughly found based on the degradation when $\lambda$ is fixed to zero.
In this way, the search range of $\rho$ can be limited, and the determined $\rho=1.5$ worked without problems even in other tasks.
However, it is true that such a manual search is burdensome, although specifying $\rho$ is easier than restricting the disturbance intensity $d^\mathrm{max}$ and LiRA can adjust the adversarial level in a state-dependent manner.
For example, the upper bound of $\rho$ could be made more user-specifiable by connecting it to acceptable ``control'' performance degradation, thus narrowing the search range beyond the current approach.
It would be desirable to achieve this connection analytically, but even if the relationship is too complex, it would be possible by introducing meta optimization methods~\citep{baratchi2024automated}.

The other important hyperparameters are $\beta$ and $\epsilon$.
$\beta$ determines the strength of the regularization of the adversary to its prior, and as $\lambda$ increases, this regularization becomes more dominant than adversarial learning, simplifying the adversary.
To make LiRA a more general technology, it would be useful to conduct a deeper analysis of the adversary's learning behavior according to $\lambda$ and $\beta$, and to design $\beta$ as a function of $\lambda$.
On the other hand, $\epsilon$ is a threshold value for the slack variable to convert inequality into equality constraints for $\lambda$ optimization.
When $\epsilon$ is small, the equality constraint is achieved through the slack variable optimization and automatic adjustment of $\lambda$ becomes slow.
In this paper, $\epsilon$ is set relatively large so that the behavior of $\lambda$ can be easily observed, but the change speed of $\lambda$ is expected to have some effects on learning.
For example, if it is too large, the learning process may become unstable because the optimization for a given $\lambda$ cannot be completed in time, or if it is too small, the sample efficiency may decrease due to the increase of similar data.
Self-paced learning~\citep{klink2021probabilistic}, in which $\epsilon$ is gradually increased to the extent that learning does not break down, may be one solution.

Although these hyperparameters have to be designed to be task-independent and generic, in real-world applications, the way of applying the disturbances and the generative model of the disturbances need to be carefully considered.
The most effective way to do this is to reflect the domain knowledge of the target, but in other cases, interrupting the agent's actions stochastically, as performed in the context of action robust RL~\citep{tessler2019action}, is a good and easy method to implement.
In this case, the generative model can be designed on the same action space as the policy, and in the non-adversarial case (i.e. $\lambda \to 1$), the agent's behavior should be passed directly to the environment.
Such a behavior is available if the adversary's prior is set to be the agent's policy, or if the interrupting probability is also included in the generative model and its prior is designed to have the minimum probability.
Note, however, that in the latter case, HRG is not applicable because the intervention signal is discrete.
Anyway, to maximize the performance of LiRA, these heuristics still need to be designed for real-world applications.

\section{Conclusion}

This study proposed a new adversarial learning framework, so-called LiRA.
LiRA enabled RL agents to improve the robustness of policy moderately while mitigating learning collapse and policy conservativeness.
To this end, adversarial learning was reformulateed using the variational inference for derivation and the light robustness for constraint.
As a result, LiRA achieved a good balance between robustness and conservativeness in comparison to the cases with/without the previous adversary, which fully prevents model learning.
In addition, LiRA was demonstrated to learn the moderately-robust quadrupedal gait controller in the real world.
The quadrupedal robot naturally acquired the ability to switch strategies, choosing to resist small disturbances in place and to step in the appropriate direction to reduce the burden when the disturbances increase.

Despite these results, there are still some issues to be addressed, as discussed in Section~\ref{sec:discussion}.
Additionally, as briefly mentioned in the main text, the disturbance-aware model can be used for MPC by introducing the domain adaptation (with the uncertainty estimator), which can contribute to reducing conservativeness.
Switching between the disturbance-marginalized and disturbance-aware models depending on the accuracy of the uncertainty estimator would allow MPC to behave robustly when the estimator is not confident, and to choose the more optimal behavior when it is confident, as like the literature~\citep{xie2022robust} (but more explicitly).
Alternatively, in the direction of improving robustness, the uncertainty of disturbance-marginalized model can be explicitly taken into account by appropriately combining a robust MPC~\citep{kohler2020computationally,zanon2020safe}.
However, as discussed in the main text, excessive robustness makes the policy conservative, so it is desirable to adaptively determine how much uncertainty to take into account like the literature~\citep{aolaritei2023wasserstein}.
After these further improvements, a comprehensive investigation with existing robust methods will be conducted to show the practical value of LiRA clearly.

\section*{Acknowledgements}

This work was supported by JST, PRESTO Grant Number JPMJPR20C3, Japan.

\appendix

\section{Auto-tuning of $\lambda$}
\label{app:lambda}

Following the literature~\citep{kobayashi2023soft}, $\lambda$ is auto-tuned as follows:
\begin{align}
    \lambda^\ast &= \arg\min_{\lambda}\mathbb{E}_{\tilde{p}_e, \tilde{r}, \pi, \varpi}
    [- \lambda \delta(s^\prime, r, s, a, d)]
    \\
    \eta^\ast &= \arg\min_{\eta}\mathbb{E}_{\tilde{p}_e, \tilde{r}, \pi, \varpi}
    [\ell^\Delta(s^\prime, r, s, a, d)]
\end{align}
where $\ell^\Delta(s^\prime, r, s, a, d)$ is given as $\epsilon$-insensitive loss.
\begin{align}
    \ell^\Delta(s^\prime, r, s, a, d) &=
    \begin{cases}
        \lambda \Delta(s; \eta) & | \delta(s^\prime, r, s, a, d) | \leq \epsilon
        \\
        | \delta(s^\prime, r, s, a, d) | & \text{otherwise}
    \end{cases}
    \label{eq:slack}
\end{align}

\section{Task configuration of worm-type robot}
\label{app:worm}

In the simulations, the task is performed with a worm-type robot placed in two-dimensional space in the forward and vertical directions with gravity.
It is simulated by Mujoco~\citep{todorov2012mujoco} under Gymnasium API and has a total length of 1~m and a mass of 10~kg.
Its linear body is segmented according to the number of joints (i.e. six for Tail2; seven for Tail3; and eight for Tail4).
The range of motion of each joint is within $\pm30$ degrees and each joint is actuated by a given torque directly.
Torques on the four joints located in the front (i.e. the head side) of the robot can be specified by the agent as actions.
On the other hand, the torques on the 2--4 joints at the rear of the robot (i.e. the tail side) are given as disturbances.
However, the maximum torque on the tail side is limited to half that of the head side.
The control frequency of this robot is 100~fps, but by repeating the same torques five times, the actual frequency for the agent and adversary is given to be 20~fps (i.e. every 0.05~s).
A total of 500 time steps (i.e. 25 seconds) is regarded to be an episode, and learning is completed after 200 episodes.

As the nominal noise, up to 20~\% of the maximum torque, which does not significantly affect the robot's behavior, is allowed.
In other words, a Gaussian distribution with $\sigma_0$ being $1/3$ of that torque is defined as the prior of the adversary.
Disturbances during learning are limited to a maximum of 20~\% (i.e. $3\sigma_0$) for the comparison method \textit{Nominal}, and a maximum of 100~\% (i.e. $15\sigma_0$) for the rest.

The task given to such a robot is to move forward at high speed.
That is, the following reward function is defined in order to the forward position of the robot tip, $x$, transits to the furthest possible $x^\prime$.
\begin{align}
    r = 100 (x^\prime - x)
\end{align}
where, 100 is multiplied so that the transition amount is in centimeters since the one in meters is too small.

In the above task configuration, the required state space is on $3 \times \#\mathrm{Joint} + 5$ dimensions:
i.e. 23 for Tail2, 26 for Tail3, and 29 for Tail4, respectively.
As mentioned above, the agent applies torque to the four joints on the head side, so the action space is four-dimensional.
Note that the disturbance itself given at that time step cannot be observed (except during adversarial learning), the disturbance before one time step can be estimated based on the dynamics of the robot, and this is also included in the state.

\section{Task configuration of quadrupedal robot}
\label{app:quad}

The quadrupedal robot used in this paper, Stella developed by Ahead.IO, is controlled internally at 100~fps by a Rasberry Pi 4.
Its target commands and disturbances are updated from RL implemented on a laptop PC (with Intel Core i7-10875H and 32GB RAM) every 5 times (i.e. 20~fps) via a local ethernet.
The reference walking trajectory is periodically generated according to the robot's internal time and the scales of the target strides $a^{\mathrm{stp}}_{xy} \in [-1, 1]$ given from RL, so that one walking step is taken in 0.2~sec.
The maximum stride length is $(s^{\mathrm{max}}_x, s^{\mathrm{max}}_y) \simeq (0.24, 0.16)$~m according to the ratio of the front-back and left-right lengths of the robot's body, and the maximum leg lift height is $0.05$~m, which is higher as the stride is larger.
However, when $a^{\mathrm{stp}}_xy$ is less than the deadzone $\pm 0.05$, the stride length is forced to be zero, and when both the forward and sideward stride lengths are zero, the robot stops in an upright position.

With the reference position of each leg given by the above walking trajectory, a virtual impedance model is introduced to generate the position commands of each leg.
Specifically, the reference position is given as the equilibrium point of the observed position with spring constants for each axis (i.e. $xyz$-axes).
The scales of the spring constants on the horizontal plane $a^{\mathrm{imp}}_{xy} \in [-1, 1]$ are given from RL, adjusting the spring constants accordingly in range $k_{xy} \in [1000, 2000]$~N/m.
The observed velocity is decayed to zero with a damper common to all axes.
This spring-damper system is disturbed by the observed force plus the virtual disturbance $d$, the range of which is $d \in [-150, 150]$~N.
Note that $d$ is added only to the legs that are ideally contact with the ground.
To satisfy this ideal ground contact as much as possible, $k_z$ is set to be as firm as possible (without oscillation) so that each leg can precisely track the reference height.

To control each leg appropriately, the target position of each leg is converted into the target joint angles for the corresponding actuators using analytical inverse kinematics.
Each joint is controlled by PD control implemented in moteus, which is the driver and controller for each actuator developed by Mjbots.
Note that $a^{\mathrm{stp}, \mathrm{imp}}_{xy}$ are actually smoothed by exponential moving average, $\bar{a}^{\mathrm{stp}, \mathrm{imp}}_{xy}$, for the above target position generation combined with the walking trajectory and the impedance model.

With such a base controller design, sufficient observation data are given to RL to make MDP valid if there is no disturbance.
Specifically, 84 dimensions are prepared in total:
the position, angular velocity, and torque (estimated from current) of each joint;
the position, velocity, and force of each leg calculated from forward kinematics and the corresponding Jacobian;
the angular velocity and acceleration measured by the IMU mounted on the robot's body;
the phase of the walking trajectory based on internal time;
and the previous smoothed actions.
However, it was difficult to solve MPC within 20~fps on such a high-dimensional observation space, while the overlapping information is expected to be included.
Therefore, as described in \ref{app:learn}, q-VAE~\citep{kobayashi2023sparse} compresses them to the 28-dimensional latent state space.
As a result, MPC is able to consider 12 time steps ahead at 20~fps, i.e. 3 walking steps ahead.

The objectives on this configuration are i) to minimize the movement of the center of the robot's body on the horizontal plane (represented by $r_1$); ii) to minimize the sum of exerted forces on the horizontal plane (represented by $r_2$); and iii) to regularize the actions (represented by $r_{3,4}$).
These aim to acquire the behaviors that if there is no disturbance, the robot stops in place, and otherwise, repels to resist it or walks to follow it.
The specific rewards are designed as follows:
\begin{align}
    \begin{split}
        r_1 &= 1 - \min\left(1, \frac{\|\sum_{l=1}^4 p_{l, xy} \|_2}{0.2}\right)
        \\
        r_2 &= 1 - \min\left(1, \frac{\sum_{l=1}^4 \|f_{l, xy}\|_2 }{100}\right)
        \\
        r_3 &= \frac{1}{10} - \frac{\| \bar{a}^{\mathrm{stp}} \|_1}{20}
        \\
        r_4 &= \frac{1}{10} - \frac{\| \bar{a}^{\mathrm{imp}} + 1 \|_1}{40}
    \end{split}
\end{align}
where $p_{l, xy}$ and $f_{l, xy}$ denote the $l$-th leg's position and force on the horizontal plane, respectively.
$0.2$ and $100$ in $r_{1,2}$ are empirically determined to avoid tipping over.
When the min operators in $r_{1,2}$ are activated to clip them, the current episode is terminated.
In addition, for safety reasons, the experimenter checks the robot from the outside and forcibly terminates it in situations where the risk of tipping over is high and/or the range of movement exceeds that limited by the cable, although this emergency stop was used only a few times during the experiment.
In these settings, a maximum of 600 steps (i.e. 30 seconds) were performed per episode, for a total of 160 episodes (i.e. 80 minutes).

\section{Learning conditions}
\label{app:learn}

The proposed framework (and also comparisons) is implemented by PyTorch~\citep{paszke2019pytorch}.
Its regularization gain to the prior is set to be $\beta=1 \times 10^{-3}$ for the scalar $\lambda$ or $\beta= 5 \times 10^{-3}$ for the state-dependent $\lambda$.
The latter is adaptively adjusted to different adversarial levels in different states and is larger than the former to allow quick disturbance suppression to match this variation.
In addition, $\rho=1.5$ is set for the gap tolerance of the prediction performance in the light-robust constraint.
For optimizing it, a replay buffer is utilized, as shown in Alg.~\ref{alg:lira}.
Its maximum capacity is 102,400 and the maximum batch size is 32.
In the learning phase, all data in the buffer are replayed uniformly at random without replacement.
In addition, a noise-robust stochastic gradient descent algorithm, AdaTerm~\citep{ilboudo2023adaterm}, is employed with its default configuration for optimizing all parameters.
AdaTerm is expected to automatically exclude outliers hidden in the worst-case data that are likely to be affected by MMB at training time.

At first, for simplicity, the following common architecture is utilized in most models.
That is, it has two fully-connected layers with 100 neurons for each.
The activation function for them are the combination of Squish function~\citep{barron2021squareplus,kobayashi2023design} and RMSNorm~\citep{zhang2019root}.

For approximating the disturbance-marginalized model $p_w(s^\prime, r \mid s, a; \theta)$, the state and action are fed into this architecture through an input layer, then parameters of a diagonal multivariate normal distribution (i.e. the location and scale parameters) are outputted.
Afterwards, $s^\prime$ and $r$ can be obtained by splitting the location parameters or sampled values into them.

By transforming it through RNF, the disturbance-aware model $p_w(s^\prime, r \mid s, a; d, \theta)$ is additionally designed.
This RNF has a small-sized architecture to output its parameters (for LRS flows~\citep{dolatabadi2020invertible}) conditioned by the disturbance: specifically, two fully-connected layers with 32 neurons for each, the activation function of which is Squaresign function~\citep{barron2021squareplus,kobayashi2023design}.
At this time, the transformability defined in RNF~\citep{kobayashi2023design} is specified as 0.99, for pursuring high expressive capability.
Anyway, after conditioning RNF with the disturbance, $s^\prime$ and $r$ generated from the disturbance-marginalized model are converted to ones from the disturbance-aware model.

For the adversary $\varpi(d \mid s; \phi)$, the common architecture first extracts the state features.
They are fed into the small-sized architecture, which outputs the parameters of CNF with LRS flows of 0.99 transformability.
When generating the disturbance, as mentioned in the main text, an uniform distribution is intorduced its base distribution.
That is, the disturbance generated from the uniform distribution is converted to one from CNF, which is adversarily optimized.

To auto-tune $\lambda$, which corresponds to the adversary level, the state-dependent slack variable $\Delta$ is also approximated with the common architecture.
Note that its boundary is satisfied by a nonlinear transformation with Squmoid function~\citep{barron2021squareplus,kobayashi2023design}.
Note that the optimization of $\Delta$ shown in eq.~\eqref{eq:slack} switches its behavior with a threshold $\epsilon=0.1$.
In addition, as described in the main text, $\lambda$ is also approximated as $\lambda(s; \zeta)$ ($\zeta$ the paramters of the common architecture) in the real-robot demonstration as well as $\Delta$ in order to obtain state-dependent adversary levels.

Finally, only in the real-robot demonstration, q-VAE~\citep{kobayashi2023sparse} is employed to compress higher-dimensional observation into low-dimensional latent state (specifically, 28 dimensions in this paper).
Specifically, an encoder with the common architecture convert the observation to the state, from which a decoder with the common architecture constructs the observation.
Both the encoder and decoder actualy output the parameters of diagonal multivariate normal distributions.
In q-VAE, they are optimized using $q$-logarithm~\citep{tsallis1988possible} with $q=0.95$, which makes the state sparse and reconstructs the obseration fairly.
The coefficients for all the terms to be minimized are set to be one for simplicity.
Note that the state extracted by the encoder is fed into the other modules after removing its computational graph, so that biases are hardly caused by factors other than compression of the observation.

\section{MPC conditions}
\label{app:mpc}

The learned model is used to obtain the optimal policy with AccelMPPI~\citep{kobayashi2022real}, a type of sampling-based MPC.
In each iteration, 1024 candidates for the simulations and 256 for the real-robot demonstration of action sequences up to 12 time steps ahead are sampled from the policy modeled by a diagonal multivariate normal distribution, and 1/4 of them, which are expected to perform well, are selected by rejecting others.
The selected candidates are evaluated by predicting the future states and rewards using the model.
According to the predicted rewards, the policy is updated to proceed to the next iteration.
Although the details of its update law is omitted in this paper (see the literature~\citep{kobayashi2022real}), the step size is 0.25, the inverse temperature is 1, the update ratio of the policy to generate negative results is 1, and the slowdown gain to adjust a accelerated gradient is 0.5.

In the simulations, the number of iterations is fixed at four times, which is mostly within the control period, to ensure reproducibility.
On the other hand, in the real-robot demonstration, iterations were sometimes terminated before four iterations to ensure real-time control.
To compensate for this lack of iteration, a warm start is employed to initialize the policy (given as a standard normal distribution at first) by moving it the step size closer to the previously optimized policy.
Note that the exploration noise added during learning is Gaussian noise with the optimized scale, since not only the mean parameter of the policy but also the scale parameter is optimized in AccelMPPI.

\bibliographystyle{elsarticle-num}
\bibliography{biblio}

\begin{thebibliography}{10}
\expandafter\ifx\csname url\endcsname\relax
  \def\url#1{\texttt{#1}}\fi
\expandafter\ifx\csname urlprefix\endcsname\relax\def\urlprefix{URL }\fi
\expandafter\ifx\csname href\endcsname\relax
  \def\href#1#2{#2} \def\path#1{#1}\fi

\bibitem{sutton2018reinforcement}
R.~S. Sutton, A.~G. Barto, Reinforcement learning: An introduction, MIT press,
  2018.

\bibitem{chua2018deep}
K.~Chua, R.~Calandra, R.~McAllister, S.~Levine, Deep reinforcement learning in
  a handful of trials using probabilistic dynamics models, in: Advances in
  Neural Information Processing Systems, 2018, pp. 4754--4765.

\bibitem{williams2018information}
G.~Williams, P.~Drews, B.~Goldfain, J.~M. Rehg, E.~A. Theodorou,
  Information-theoretic model predictive control: Theory and applications to
  autonomous driving, IEEE Transactions on Robotics 34~(6) (2018) 1603--1622.

\bibitem{nagabandi2020deep}
A.~Nagabandi, K.~Konolige, S.~Levine, V.~Kumar, Deep dynamics models for
  learning dexterous manipulation, in: Conference on Robot Learning, PMLR,
  2020, pp. 1101--1112.

\bibitem{luque2024model}
A.~Luque, D.~Parent, A.~Colom{\'e}, C.~Ocampo-Martinez, C.~Torras, Model
  predictive control for dynamic cloth manipulation: Parameter learning and
  experimental validation, IEEE Transactions on Control Systems Technology
  (2024).

\bibitem{hachimine2023learning}
T.~Hachimine, J.~Morimoto, T.~Matsubara, Learning to shape by grinding:
  Cutting-surface-aware model-based reinforcement learning, IEEE Robotics and
  Automation Letters (2023).

\bibitem{yang2020data}
Y.~Yang, K.~Caluwaerts, A.~Iscen, T.~Zhang, J.~Tan, V.~Sindhwani, Data
  efficient reinforcement learning for legged robots, in: Conference on Robot
  Learning, PMLR, 2020, pp. 1--10.

\bibitem{kuo2023reinforcement}
C.-Y. Kuo, H.~Shin, T.~Matsubara, Reinforcement learning with energy-exchange
  dynamics for spring-loaded biped robot walking, IEEE Robotics and Automation
  Letters (2023).

\bibitem{lambert2019low}
N.~O. Lambert, D.~S. Drew, J.~Yaconelli, S.~Levine, R.~Calandra, K.~S. Pister,
  Low-level control of a quadrotor with deep model-based reinforcement
  learning, IEEE Robotics and Automation Letters 4~(4) (2019) 4224--4230.

\bibitem{arcari2023bayesian}
E.~Arcari, M.~V. Minniti, A.~Scampicchio, A.~Carron, F.~Farshidian, M.~Hutter,
  M.~N. Zeilinger, Bayesian multi-task learning mpc for robotic mobile
  manipulation, IEEE Robotics and Automation Letters (2023).

\bibitem{zhang2021reinforcement}
T.~Zhang, H.~Mo, Reinforcement learning for robot research: A comprehensive
  review and open issues, International Journal of Advanced Robotic Systems
  18~(3) (2021) 17298814211007305.

\bibitem{morimoto2005robust}
J.~Morimoto, K.~Doya, Robust reinforcement learning, Neural computation 17~(2)
  (2005) 335--359.

\bibitem{kingma2014auto}
D.~P. Kingma, M.~Welling, Auto-encoding variational bayes, in: International
  Conference on Learning Representations, 2014.

\bibitem{fischetti2009light}
M.~Fischetti, M.~Monaci, Light robustness, Robust and online large-scale
  optimization: Models and techniques for transportation systems (2009) 61--84.

\bibitem{kobayashi2023soft}
T.~Kobayashi, Soft actor-critic algorithm with truly-satisfied inequality
  constraint, arXiv preprint arXiv:2303.04356 (2023).

\bibitem{todorov2012mujoco}
E.~Todorov, T.~Erez, Y.~Tassa, Mujoco: A physics engine for model-based
  control, in: IEEE/RSJ international conference on intelligent robots and
  systems, IEEE, 2012, pp. 5026--5033.

\bibitem{kumar2010self}
M.~Kumar, B.~Packer, D.~Koller, Self-paced learning for latent variable models,
  Advances in neural information processing systems 23 (2010).

\bibitem{klink2020self}
P.~Klink, C.~D'Eramo, J.~R. Peters, J.~Pajarinen, Self-paced deep reinforcement
  learning, Advances in Neural Information Processing Systems 33 (2020)
  9216--9227.

\bibitem{hoeller2024anymal}
D.~Hoeller, N.~Rudin, D.~Sako, M.~Hutter, Anymal parkour: Learning agile
  navigation for quadrupedal robots, Science Robotics 9~(88) (2024) eadi7566.

\bibitem{radosavovic2024real}
I.~Radosavovic, T.~Xiao, B.~Zhang, T.~Darrell, J.~Malik, K.~Sreenath,
  Real-world humanoid locomotion with reinforcement learning, Science Robotics
  9~(89) (2024) eadi9579.

\bibitem{tobin2017domain}
J.~Tobin, R.~Fong, A.~Ray, J.~Schneider, W.~Zaremba, P.~Abbeel, Domain
  randomization for transferring deep neural networks from simulation to the
  real world, in: IEEE/RSJ international conference on intelligent robots and
  systems, IEEE, 2017, pp. 23--30.

\bibitem{ramos2019bayessim}
F.~Ramos, R.~C. Possas, D.~Fox, Bayessim: adaptive domain randomization via
  probabilistic inference for robotics simulators, in: Robotics: Science and
  Systems, 2019.

\bibitem{muratore2022neural}
F.~Muratore, T.~Gruner, F.~Wiese, B.~Belousov, M.~Gienger, J.~Peters, Neural
  posterior domain randomization, in: Conference on Robot Learning, PMLR, 2022,
  pp. 1532--1542.

\bibitem{andrychowicz2020learning}
O.~M. Andrychowicz, B.~Baker, M.~Chociej, R.~Jozefowicz, B.~McGrew,
  J.~Pachocki, A.~Petron, M.~Plappert, G.~Powell, A.~Ray, et~al., Learning
  dexterous in-hand manipulation, The International Journal of Robotics
  Research 39~(1) (2020) 3--20.

\bibitem{rudin2022learning}
N.~Rudin, D.~Hoeller, P.~Reist, M.~Hutter, Learning to walk in minutes using
  massively parallel deep reinforcement learning, in: Conference on Robot
  Learning, PMLR, 2022, pp. 91--100.

\bibitem{chen2022understanding}
X.~Chen, J.~Hu, C.~Jin, L.~Li, L.~Wang, Understanding domain randomization for
  sim-to-real transfer, in: International Conference on Learning
  Representations, 2022.

\bibitem{pinto2017robust}
L.~Pinto, J.~Davidson, R.~Sukthankar, A.~Gupta, Robust adversarial
  reinforcement learning, in: International Conference on Machine Learning,
  PMLR, 2017, pp. 2817--2826.

\bibitem{gleave2020adversarial}
A.~Gleave, M.~Dennis, C.~Wild, N.~Kant, S.~Levine, S.~Russell, Adversarial
  policies: Attacking deep reinforcement learning, in: International Conference
  on Learning Representations, 2020.

\bibitem{zhai2022robust}
P.~Zhai, J.~Luo, Z.~Dong, L.~Zhang, S.~Wang, D.~Yang, Robust adversarial
  reinforcement learning with dissipation inequation constraint, in: AAAI
  Conference on Artificial Intelligence, Vol.~36, 2022, pp. 5431--5439.

\bibitem{srivastava2017veegan}
A.~Srivastava, L.~Valkov, C.~Russell, M.~U. Gutmann, C.~Sutton, Veegan:
  Reducing mode collapse in gans using implicit variational learning, Advances
  in neural information processing systems 30 (2017).

\bibitem{liu2019spectral}
K.~Liu, W.~Tang, F.~Zhou, G.~Qiu, Spectral regularization for combating mode
  collapse in gans, in: IEEE/CVF international conference on computer vision,
  2019, pp. 6382--6390.

\bibitem{huang2022robust}
P.~Huang, M.~Xu, F.~Fang, D.~Zhao, Robust reinforcement learning as a
  stackelberg game via adaptively-regularized adversarial training, in:
  International Joint Conference on Artificial Intelligence, 2022.

\bibitem{petrik2019beyond}
M.~Petrik, R.~H. Russel, Beyond confidence regions: Tight bayesian ambiguity
  sets for robust mdps, Advances in neural information processing systems 32
  (2019).

\bibitem{lechner2023revisiting}
M.~Lechner, A.~Amini, D.~Rus, T.~A. Henzinger, Revisiting the adversarial
  robustness-accuracy tradeoff in robot learning, IEEE Robotics and Automation
  Letters 8~(3) (2023) 1595--1602.

\bibitem{huang2023trade}
J.~Huang, H.~J. Choi, N.~Figueroa, Trade-off between robustness and rewards
  adversarial training for deep reinforcement learning under large
  perturbations, IEEE Robotics and Automation Letters 8~(12) (2023) 8018--8025.

\bibitem{yu2017preparing}
W.~Yu, J.~Tan, C.~K. Liu, G.~Turk, Preparing for the unknown: Learning a
  universal policy with online system identification, in: Robotics: Science and
  Systems, 2017.

\bibitem{semage2022uncertainty}
B.~L. Semage, T.~G. Karimpanal, S.~Rana, S.~Venkatesh, Uncertainty aware system
  identification with universal policies, in: International Conference on
  Pattern Recognition, IEEE, 2022, pp. 2321--2327.

\bibitem{ilboudo2023domains}
W.~E.~L. Ilboudo, T.~Kobayashi, T.~Matsubara, Domains as objectives:
  Multi-domain reinforcement learning with convex-coverage set learning for
  domain uncertainty awareness, in: IEEE/RSJ International Conference on
  Intelligent Robots and Systems, IEEE, 2023, pp. 5622--5629.

\bibitem{kobayashi2022real}
T.~Kobayashi, K.~Fukumoto, Real-time sampling-based model predictive control
  based on reverse kullback-leibler divergence and its adaptive acceleration,
  arXiv preprint arXiv:2212.04298 (2022).

\bibitem{lenz2015deepmpc}
I.~Lenz, R.~A. Knepper, A.~Saxena, Deepmpc: Learning deep latent features for
  model predictive control., in: Robotics: Science and Systems, Vol.~10, Rome,
  Italy, 2015.

\bibitem{kobayashi2023sparse}
T.~Kobayashi, R.~Watanuki, Sparse representation learning with modified q-vae
  towards minimal realization of world model, Advanced Robotics 37~(13) (2023)
  807--827.

\bibitem{hayes2022practical}
C.~F. Hayes, R.~R{\u{a}}dulescu, E.~Bargiacchi, J.~K{\"a}llstr{\"o}m,
  M.~Macfarlane, M.~Reymond, T.~Verstraeten, L.~M. Zintgraf, R.~Dazeley,
  F.~Heintz, et~al., A practical guide to multi-objective reinforcement
  learning and planning, Autonomous Agents and Multi-Agent Systems 36~(1)
  (2022) 26.

\bibitem{kohler2020computationally}
J.~K{\"o}hler, R.~Soloperto, M.~A. M{\"u}ller, F.~Allg{\"o}wer, A
  computationally efficient robust model predictive control framework for
  uncertain nonlinear systems, IEEE Transactions on Automatic Control 66~(2)
  (2020) 794--801.

\bibitem{zanon2020safe}
M.~Zanon, S.~Gros, Safe reinforcement learning using robust mpc, IEEE
  Transactions on Automatic Control 66~(8) (2020) 3638--3652.

\bibitem{aotani2024cooperative}
T.~Aotani, T.~Kobayashi, Cooperative transport by manipulators with
  uncertainty-aware model-based reinforcement learning, in: IEEE/SICE
  International Symposium on System Integration, IEEE, 2024, pp. 959--964.

\bibitem{kobayashi2023design}
T.~Kobayashi, T.~Aotani, Design of restricted normalizing flow towards
  arbitrary stochastic policy with computational efficiency, Advanced Robotics
  37~(12) (2023) 719--736.

\bibitem{ganin2016domain}
Y.~Ganin, E.~Ustinova, H.~Ajakan, P.~Germain, H.~Larochelle, F.~Laviolette,
  M.~March, V.~Lempitsky, Domain-adversarial training of neural networks,
  Journal of machine learning research 17~(59) (2016) 1--35.

\bibitem{higgins2017beta}
I.~Higgins, L.~Matthey, A.~Pal, C.~P. Burgess, X.~Glorot, M.~M. Botvinick,
  S.~Mohamed, A.~Lerchner, beta-vae: Learning basic visual concepts with a
  constrained variational framework., in: International Conference on Learning
  Representations, 2017.

\bibitem{papamakarios2021normalizing}
G.~Papamakarios, E.~Nalisnick, D.~J. Rezende, S.~Mohamed, B.~Lakshminarayanan,
  Normalizing flows for probabilistic modeling and inference, Journal of
  Machine Learning Research 22~(57) (2021) 1--64.

\bibitem{williams1992simple}
R.~J. Williams, Simple statistical gradient-following algorithms for
  connectionist reinforcement learning, Machine learning 8 (1992) 229--256.

\bibitem{mohaghegh2020advflow}
H.~Mohaghegh~Dolatabadi, S.~Erfani, C.~Leckie, Advflow: Inconspicuous black-box
  adversarial attacks using normalizing flows, Advances in Neural Information
  Processing Systems 33 (2020) 15871--15884.

\bibitem{haarnoja2018soft}
T.~Haarnoja, A.~Zhou, P.~Abbeel, S.~Levine, Soft actor-critic: Off-policy
  maximum entropy deep reinforcement learning with a stochastic actor, in:
  International conference on machine learning, PMLR, 2018, pp. 1861--1870.

\bibitem{aotani2021meta}
T.~Aotani, T.~Kobayashi, K.~Sugimoto, Meta-optimization of bias-variance
  trade-off in stochastic model learning, IEEE Access 9 (2021) 148783--148799.

\bibitem{dodge2003oxford}
Y.~Dodge, The Oxford dictionary of statistical terms, Oxford University Press,
  USA, 2003.

\bibitem{agarwal2021deep}
R.~Agarwal, M.~Schwarzer, P.~S. Castro, A.~C. Courville, M.~Bellemare, Deep
  reinforcement learning at the edge of the statistical precipice, Advances in
  neural information processing systems 34 (2021) 29304--29320.

\bibitem{shi2024rethinking}
F.~Shi, C.~Zhang, T.~Miki, J.~Lee, M.~Hutter, S.~Coros, Rethinking robustness
  assessment: Adversarial attacks on learning-based quadrupedal locomotion
  controllers, arXiv preprint arXiv:2405.12424 (2024).

\bibitem{mcinnes2018umap}
L.~McInnes, J.~Healy, N.~Saul, L.~Grossberger, Umap: Uniform manifold
  approximation and projection, The Journal of Open Source Software 3~(29)
  (2018) 861.

\bibitem{kobayashi2022l2c2}
T.~Kobayashi, {L2C2}: Locally lipschitz continuous constraint towards stable
  and smooth reinforcement learning, in: IEEE/RSJ International Conference on
  Intelligent Robots and Systems, IEEE, 2022, pp. 4032--4039.

\bibitem{wenzel2020hyperparameter}
F.~Wenzel, J.~Snoek, D.~Tran, R.~Jenatton, Hyperparameter ensembles for
  robustness and uncertainty quantification, Advances in Neural Information
  Processing Systems 33 (2020) 6514--6527.

\bibitem{wilson2021balancing}
R.~C. Wilson, E.~Bonawitz, V.~D. Costa, R.~B. Ebitz, Balancing exploration and
  exploitation with information and randomization, Current opinion in
  behavioral sciences 38 (2021) 49--56.

\bibitem{seo2021state}
Y.~Seo, L.~Chen, J.~Shin, H.~Lee, P.~Abbeel, K.~Lee, State entropy maximization
  with random encoders for efficient exploration, in: International Conference
  on Machine Learning, PMLR, 2021, pp. 9443--9454.

\bibitem{oh2022model}
Y.~Oh, J.~Shin, E.~Yang, S.~J. Hwang, Model-augmented prioritized experience
  replay, in: International Conference on Learning Representations, 2022.

\bibitem{pan2022understanding}
Y.~Pan, J.~Mei, A.-m. Farahmand, M.~White, H.~Yao, M.~Rohani, J.~Luo,
  Understanding and mitigating the limitations of prioritized experience
  replay, in: Uncertainty in Artificial Intelligence, PMLR, 2022, pp.
  1561--1571.

\bibitem{lin2024smooth}
X.~Lin, X.~Zhang, Z.~Yang, F.~Liu, Z.~Wang, Q.~Zhang, Smooth tchebycheff
  scalarization for multi-objective optimization, in: International Conference
  on Machine Learning, PMLR, 2024, pp. 30479--30509.

\bibitem{baratchi2024automated}
M.~Baratchi, C.~Wang, S.~Limmer, J.~N. van Rijn, H.~Hoos, T.~B{\"a}ck,
  M.~Olhofer, Automated machine learning: past, present and future, Artificial
  intelligence review 57~(5) (2024) 122.

\bibitem{klink2021probabilistic}
P.~Klink, H.~Abdulsamad, B.~Belousov, C.~D'Eramo, J.~Peters, J.~Pajarinen, A
  probabilistic interpretation of self-paced learning with applications to
  reinforcement learning, Journal of Machine Learning Research 22~(182) (2021)
  1--52.

\bibitem{tessler2019action}
C.~Tessler, Y.~Efroni, S.~Mannor, Action robust reinforcement learning and
  applications in continuous control, in: International Conference on Machine
  Learning, PMLR, 2019, pp. 6215--6224.

\bibitem{xie2022robust}
A.~Xie, S.~Sodhani, C.~Finn, J.~Pineau, A.~Zhang, Robust policy learning over
  multiple uncertainty sets, in: International Conference on Machine Learning,
  PMLR, 2022, pp. 24414--24429.

\bibitem{aolaritei2023wasserstein}
L.~Aolaritei, M.~Fochesato, J.~Lygeros, F.~D{\"o}rfler, Wasserstein tube mpc
  with exact uncertainty propagation, in: IEEE Conference on Decision and
  Control, IEEE, 2023, pp. 2036--2041.

\bibitem{paszke2019pytorch}
A.~Paszke, S.~Gross, F.~Massa, A.~Lerer, J.~Bradbury, G.~Chanan, T.~Killeen,
  Z.~Lin, N.~Gimelshein, L.~Antiga, et~al., Pytorch: An imperative style,
  high-performance deep learning library, Advances in neural information
  processing systems 32 (2019).

\bibitem{ilboudo2023adaterm}
W.~E.~L. Ilboudo, T.~Kobayashi, T.~Matsubara, Adaterm: Adaptive t-distribution
  estimated robust moments for noise-robust stochastic gradient optimization,
  Neurocomputing 557 (2023) 126692.

\bibitem{barron2021squareplus}
J.~T. Barron, Squareplus: A softplus-like algebraic rectifier, arXiv preprint
  arXiv:2112.11687 (2021).

\bibitem{zhang2019root}
B.~Zhang, R.~Sennrich, Root mean square layer normalization, Advances in Neural
  Information Processing Systems 32 (2019).

\bibitem{dolatabadi2020invertible}
H.~M. Dolatabadi, S.~Erfani, C.~Leckie, Invertible generative modeling using
  linear rational splines, in: International Conference on Artificial
  Intelligence and Statistics, PMLR, 2020, pp. 4236--4246.

\bibitem{tsallis1988possible}
C.~Tsallis, Possible generalization of boltzmann-gibbs statistics, Journal of
  statistical physics 52~(1-2) (1988) 479--487.

\end{thebibliography}


\end{document}